\PassOptionsToPackage{usenames,x11names,dvipsnames,svgnames}{xcolor}
\documentclass[ai,article,accept,moreauthors,pdftex]{Definitions/mdpi-arxiv}

\firstpage{150} 
\pubvolume{2}
\issuenum{1}
\articlenumber{0}
\pubyear{2021}
\copyrightyear{2021}
\datereceived{15 March 2021} 
\dateaccepted{7 April 2021} 
\datepublished{12 April 2021} 
\hreflink{https://doi.org/10.3390/ai2020010} 

\usepackage{calc,xspace}
\usepackage{booktabs}
\usepackage{multirow}
\usepackage{textcomp}
\usepackage{nicefrac}
\usepackage{microtype}
\usepackage{mathtools}
\usepackage{subcaption}
\usepackage[euler]{textgreek}
\usepackage[ruled,linesnumbered]{algorithm2e}
\usepackage{pgfplots}
\pgfplotsset{compat=1.16}
\usepackage{tikz}
\usetikzlibrary{shapes,arrows,chains,pgfplots.dateplot}

\newcommand{\eg}{e.g.,\xspace}
\newcommand{\ie}{i.e.,\xspace}
\newcommand{\termdefn}[1]{\emph{#1}}

\definecolor{darkcoral}{rgb}{0.8, 0.36, 0.27}

\newcommand{\heur}[1]{\textit{#1}}
\newcommand{\policy}[1]{\textsf{#1}}
\newcommand{\param}[1]{\textit{#1}}

\Title{A Study of Learning Search Approximation in Mixed Integer Branch and Bound: Node Selection in SCIP}
\TitleCitation{A Study of Learning Search Approximation in Mixed Integer Branch and Bound: Node Selection in SCIP}

\Author{%
   Kaan Yilmaz and Neil Yorke-Smith *\orcidC{}
}   
   
\AuthorNames{Kaan Yilmaz
and Neil Yorke-Smith}

\AuthorCitation{Yilmaz, K.;
Yorke-Smith, N.}

\address[1]{%
Faculty of Electrical Engineering, Mathematics and Computer Science, Delft University of Technology,
2600 GA, Delft, The Netherlands.
 \textbf{NOTE this is an authors' preprint, not the publisher's version of the article.} }

\corres{\hangafter=1 \hangindent=1.05em \hspace{-0.82em}Correspondence: n.yorke-smith@tudelft.nl}

\abstract{%
In line with the growing trend of using machine learning to help solve combinatorial optimisation problems, one promising idea is to improve node selection within a mixed integer programming (MIP) branch-and-bound tree by using a learned policy.
 {Previous work using imitation learning indicates the feasibility of acquiring a node selection policy, by learning an adaptive node searching order.
In contrast,} our imitation learning policy is focused solely on learning which of a node's children to select.
We present an offline method to learn such a policy in two settings: one that comprises a heuristic by committing to pruning of nodes; one that is exact and backtracks from a leaf to guarantee finding the optimal integer solution.  The former setting corresponds to a child selector during plunging, while the latter is akin to a diving heuristic.
We apply the policy within the popular open-source solver SCIP, in both heuristic and exact settings.
Empirical results on five MIP datasets indicate that our node selection policy leads to solutions significantly more quickly than the state-of-the-art precedent in the literature.
While we do not beat the highly-optimised SCIP state-of-practice baseline node selector in terms of solving time on exact solutions, our heuristic policies have a consistently better optimality gap than all baselines, if the accuracy of the predictive model is sufficient.
Further, the results also indicate that, when a time limit is applied, our heuristic method finds better solutions than all baselines in the majority of problems tested.
We explain the results by showing that the learned policies have imitated the SCIP baseline, but without the latter's early plunge abort.  Our recommendation is that, despite the clear improvements over the literature, this kind of MIP child selector is better seen in a broader approach to using learning in MIP branch-and-bound tree decisions.
}

\keyword{mixed integer programming; node selection; machine learning; approximate pruning; imitation learning; SCIP}

\begin{document}

\section{Introduction}
\label{sec:intro}
Hard constrained optimisation problems (COPs) exist in many different applications.  Examples include airline scheduling \citep{DBLP:journals/anor/BaylissMAP17} and CPU efficiency maximisation \citep{lombardi2017empirical}.
Perhaps the most common approach for modelling and solving COPs in practice is mixed integer linear programming (MILP, or simply MIP).  State-of-the-art MIP solvers perform sophisticated pre-solve mechanisms followed by branch-and-bound search with cuts and additional heuristics \citep{scip6}.

\textls[-30]{A growing trend is to use machine learning (ML) to improve COP solving.  \mbox{Bengio et al. \citep{DBLP:journals/eor/BengioLP21}}} survey the potential of ML to assist MIP solvers.  One promising idea is to improve node selection within a MIP branch-and-bound tree by using a learned policy.
A policy is a function that maps states to actions, where in this context an action is the next node to select.  
However, research in ML-based node selection is scarce, as the only available literature is the work of \mbox{He et al. \citep{He2014:learning}}.
The objective of this article is to explore learning approximate node selection policies.

Node selection rules are important because the solver must find the balance between exploring nodes with good lower bounds, which tend to be found at the top of the branch-and-bound search tree, and finding feasible solutions fast, which often happens deeper into the tree.  The former task allows for the global lower bound to improve (which allows us to prove optimality), whereas the latter is the main driver for pruning unnecessary nodes.

This article contributes a novel approach to MIP node selection by using an offline learned policy.  
We obtain a node selection and pruning policy with imitation learning, \mbox{a type} of supervised learning.
In contrast to \mbox{He et al. \citep{He2014:learning}}, our policy learns solely to choose which of a node's children it should select.  This encourages finding solutions quickly, \mbox{as opposed} to learning a breadth-first search-like policy.  Further, we generalise the expert demonstration process by sampling paths that lead to the best $k$ solutions, instead of only the top single solution.  The motivation is to obtain different state-action pairs that lead to good solutions compared to only using the top solution, in order to aid learning within a deep learning context.

We study two settings: the first is heuristic that approximates  {the solution process} by committing to pruning of nodes.  In this way, the solver might find good or optimal solutions more quickly, however with the possibility of overlooking optimal solutions.  \mbox{This first} setting has similarities to a diving heuristic.  By contrast, the second setting is exact: when reaching a leaf the solver backtracks up the branch-and-bound tree to use a different path.
In the first setting, the learned policy is used as an heuristic method.
\mbox{This is} akin to a diving heuristic, with the difference that it uses the default branching rule as a variable fixing strategy.
In the second setting, the learned policy is used as a child selector during plunging.  The potential of learning here is to augment node selection rules, \mbox{such that} the child selection process is better informed than the simple heuristics that solvers typically use.
 
We apply the learned policy within the popular open-source solver SCIP \citep{scip6}, in both settings.  The results indicate that, while our node selector finds (optimal) solutions a little slower on average than the current default SCIP node selector,
\heur{BestEstimate}, it does so much more quickly than the state-of-the-art in ML-based node selectors.
Moreover,
\mbox{our heuristic} method finds better initial solutions than \heur{BestEstimate}, albeit in a higher solving time.  Overall, our heuristic policies have a consistently better optimality gap than all baselines if the accuracy of the predictive model adds value to prediction.  Further, when a time limit is applied, our heuristic method finds better solutions than all the baselines (including \heur{BestEstimate}) in three of five problem classes tested and ties in a fourth \mbox{problem class}.

While these results indicate the potential of ML in MIP node selection, we uncover several limitations.  This analysis comes from showing that the learned policies have imitated the SCIP baseline, but without the latter's early plunge abort.
Therefore our recommendation is that, despite the clear improvements over the literature, this kind of MIP child node selector is better seen in a broader approach to using learning in MIP branch-and-bound tree decisions.

The outline of this article is as follows: Section~\ref{sec:bg} contains preliminaries, Section~\ref{sec:method} specifies the imitation learning approach to node selection and pruning, Section~\ref{sec:results} reports the results on benchmark MIP instances (with additional instances reported in the Appendix~A),
Section~\ref{sec:discuss} discusses the results, Section~\ref{sec:rw} reviews related work, and Section~\ref{sec:conc} concludes with future directions.

\section{Background}
\label{sec:bg}
\begin{algorithm}[!h]
    \SetAlgoLined
    \SetKwInOut{Input}{input}\SetKwInOut{Output}{output}
    \Input{Root $R$, which is a node representing the original problem}
    \Output{Optimal solution if one exists}
    \BlankLine
    
    \emph{$R$\text{.dualBound}} $\leftarrow$ $-\infty$ \tcp{Initialise dual bound}
    \emph{$PQ$} $\leftarrow$ \{\emph{$R$}\} ~~~~~~~~~~~~~~~~~\tcp{Node priority queue}
    \emph{$B_P$} $\leftarrow$ $\infty$ ~~~~~~~~~~~~~~~~~~~~\tcp{Primal bound}
    \emph{$S^*$} $\leftarrow$ null ~~~~~~~~~~~~~~~~~\tcp{Optimal solution}
    \BlankLine
    \While{\emph{$PQ$} \text{is not empty}}{
        \tcp{Remove and return the node in front of the queue}
        \emph{$N$} $\leftarrow$ \emph{$PQ$}.poll()\; 
        \BlankLine
        
        \If{\emph{$N$\text{.dualBound}} $ \ge B_P$}{
            \tcp{Parent of N had a relaxed solution worse than current best integer feasible solution, skip solving relaxation and prune}
            \textbf{continue}
        }
        
        \emph{$S_r$} $\leftarrow$ solveRelaxation($N$)\;
        
        \If{\emph{$S_r$} \text{is not feasible}}{
            \tcp{Infeasible relaxation can not lead to a feasible solution for the original problem}
            \textbf{continue}\;
        }
        
        \emph{$O_r$} $\leftarrow$ \emph{$S_r$}\text{.objectiveValue}\;
        \BlankLine
        \If{\emph{$O_r$} $>$ \emph{$B_P$}}{
            \tcp{This subtree cannot contain any solution better than the current best (pruning)}
            \textbf{continue}\;
        }
        \If{\emph{$S_r$} \text{is integer feasible}}{
            \emph{$B_P$} $\leftarrow$ \emph{$O_r$}\;
            \emph{$S^*$} $\leftarrow$ \emph{$S_r$}\;
            \tcp{Found incumbent solution no worse than current best}
            \textbf{continue}\;
        }
        \emph{$V$} $\leftarrow$ variableSelection(\emph{$S_r$})\;
        \emph{$a$} $\leftarrow$ floor(\emph{$V$}\text{.value})\;
        \emph{$L$} $\leftarrow$ copyAndAddConstraint(\emph{$N$}, \emph{$V$} $\le$ \emph{$a$})\;
        \emph{$R$} $\leftarrow$ copyAndAddConstraint(\emph{$N$}, \emph{$V$} $\ge$ \emph{$a$} $+ 1$)\;
        \emph{$L$\text{.dualBound}} $\leftarrow$ $O_r$\;
        \emph{$R$\text{.dualBound}} $\leftarrow$ $O_r$\;
        \emph{$PQ$}.add($L$)\;
        \emph{$PQ$}.add($R$)\;
    }
    \Return \emph{$S^*$}\;
    \caption{Solve a minimisation MIP problem using branch and bound.}
    \label{algo_disjdecomp}
\end{algorithm}

Mixed integer programming (MIP) is a familiar approach to constraint optimisation problems.
A MIP requires one or more variables to decide on, a set of linear constraints that need to be met, and a linear objective function, which produces an objective value 
without loss of generality to be minimised.
Thus we have, for $y$ the objective value and $x$ the vector of decision variables to decide on:

\begin{equation}
    \label{MILP-formulation}
        \min y \text{ := } c^Tx
        \text{ s.t. }  Ax \ge b\cdot
        x\in \mathbb{Z}^k \times \mathbb{R}^{n-k}, k > 0
\end{equation}

\noindent
where $A$ is an $m$ $\times$ $n$ constraint matrix with $m$ constraints and $n$ variables; $c$ is a \mbox{$n$ $\times$ $1$ vector}.

At \mbox{least one variable has integer domain in a MIP; if all variables have} continuous domains then the problem is a linear program (LP). 
Since general MIP problems cannot be solved in polynomial time, 
the LP relaxation of \eqref{MILP-formulation} relaxes the integer constraints.
A series of LP relaxation can be leveraged in the MIP solving process.  For minimisation problems, the solution of the relaxation provides a lower bound on the original MIP problem.

Equation~(\ref{MILP-formulation}) is also referred to as the primal problem.  The primal bound is the objective value of a solution that is feasible, but not necessarily optimal.
This is referred to as a `pessimistic' bound.  The dual bound is the objective value of the solution of an LP relaxation, which is not necessarily feasible.  This is referred to as an `optimistic' bound.
The \termdefn{integrality gap} is defined as:

\begin{equation}
    I_G =
    \begin{cases}
    \frac{|B_P - B_D|}{\min(|B_P|,|B_D|)|},& \text{if } \text{sign}(B_P) = \text{sign}(B_D) \\
    \infty,              & \text{otherwise}
    \end{cases}
\end{equation}

\noindent
where $B_P$ is the primal bound, $B_D$ is the dual bound, and $\text{sign}(\cdot)$ returns the sign of its argument.  Note this definition is that used by the SCIP solver \citep{scip6}, and requires care to handle the infinite gaps case.

Related to the integrality gap is the \emph{optimality gap}: the difference between the objective function value of the best found solution and that of the optimal solution.
Both the integrality gap and the optimality gap are monotonically reduced during the solving process.  The solving process combines inference, notably in the form of inferred constraints (`cuts'), and search, usually in a branch-and-bound framework.

\label{background:branch-and-bound}

Branch and bound \citep{landautomatic} is the most common constructive search approach to solving MIP problems.
The state space of possible solutions is explored with a growing tree.  \mbox{The root} node consists of all solutions.  At every node, an unassigned integer variable is chosen to branch on.  Every node has two children: candidate solutions for the lower and upper bound respectively of the chosen variable.
Note that a node (and its entire sub-tree) is pruned when the solution of the relaxation at that node is worse than the current primal bound.
The main steps of a standard MIP branch-and-bound algorithm are given in Algorithm~\ref{algo_disjdecomp}.

Choosing on which variable to branch
is not trivial and affects the size of the resulting search tree, and therefore the time to find solutions and prove optimality.  
For example, \mbox{in the} SCIP solver,
the default variable selection heuristic (the `brancher') is hybrid branching \citep{achterberg2009hybrid}.  Other variable selection heuristics are for example pseudo-cost branching \citep{DBLP:journals/mp/BenichouGGHRV71} and reliability branching on pseudo-cost values \cite{achterberg2005branching}.  The brancher can inform the node selector which child it prefers; it is up to the node selector, however, to choose the child.  The child preferred by the brancher, if any, is called the priority child.

SCIP's brancher prefers the priority child by using the \emph{PrioChild} property.  It is used as a feature in our work.
The intuition is to maximise the number of inferences and pushing the solution away from the root node values.
In more detail, the left child priority value is calculated by SCIP as:
    $P_L = I_L (V_r - V + 1)$
and the right child priority value as:
    $P_R = I_R (V - V_r + 1)$
where $I_L$ (respectively $I_R$) is the average number of inferences at the left child (right child), $V_r$ is the value of the relaxation of the branched variable at the root node and $V$ is the value of the relaxation of the branched variable at the current node.  \mbox{An inference} is defined as a deduction of another variable after tightening the bound on a branched variable \citep{achterberg2007constraint}. If $P_L > P_R$, then the left child is prioritised over the right child, if $P_L < P_R$, then the right child is prioritised.  If they are equal, then none are prioritised. Note that while this rule for priority does not necessarily hold for all branchers in general, it does hold for the standard SCIP brancher.

Choosing on which node to prioritise for exploration over another node is defined by the node selector.
As is the case for branching, different heuristics exist for node selection.  Among these are depth-first search (\heur{DFS}), breadth-first search (\heur{BFS}), \heur{RestartDFS} (restarting \heur{DFS} at the
best bound node
after a fixed amount of newly-explored nodes) and \heur{BestEstimate}.  The latter is the default node selector in the SCIP solver from version~6.  It uses an estimate of the objective function at a node to select the next node and it prefers diving deep into the search tree
(see \url{www.scipopt.org/doc/html/nodesel\_\_estimate\_8c\_source.php}, accessed 01-Sep-20).

In more detail, \heur{BestEstimate} 
assigns a score to each node, which takes into account the quality of its dual bound, plus an objective value penalty for lack of integrality.  \mbox{In particular}, the penalty is calculated using the variable's pseudo-costs, which act as an indicator of the per unit objective value increase for shifting a variable downwards or upwards.  SCIP actually uses, \heur{BestEstimate} with plunging, which applies depth-first search (selecting children/sibling nodes) until this is no longer possible or until a diving abort mechanism is triggered.  SCIP then chooses a node according to the \heur{BestEstimate} score.

Node selection heuristics can be grouped into two general strategies \citep{achterberg2008constraint}.  The first is choosing the node with the best lower bound in order to increase the global dual bound.  The second strategy is diving into the tree to search for feasible solutions and decrease the primal bound.  This has the advantage to prune more nodes and decrease the search space.  In this article we use the second strategy to develop a novel heuristic using machine learning, leveraging local variable, local node and global tree features, in order to predict as far as possible the best possible child node to be selected.

We can further leverage node pruning to create a heuristic algorithm.  The goal is then to prune nodes that lead to bad solutions.  Correctly pruning sub-trees that do not contain an optimal solution is analogous to taking the shortest path to an optimal solution, \mbox{which obviously} minimises the solving time.  It is generally preferred to find feasible solutions quickly, as this enables the node pruner to prune more sub-trees (due to bounding), \mbox{with the} effect of decreasing the search space. 
There is no guarantee, however, that the optimal solution is not pruned.

\section{Approach}
\label{sec:method}

Recall that our goal is to obtain a MIP node selection policy using machine learning, and to use it in a MIP solver.  The policy should lead to promising solutions more quickly in the branch-and-bound tree, while pruning as few good solutions as possible.

Our approach is to obtain a node selector
by imitation learning.  A policy maps a state $s_t$ to an action $a_t$.  In our case $s_t$ consists of features gathered within the branch-and-bound process.  
The features consist of branched variable features, node features and global features.  The branched variable features are derived from \mbox{Gasse et al.\@ \citep{DBLP:conf/nips/GasseCFC019}}.  See Table~\ref{tab:features} 
for the list of features.  Note that we define a separate \emph{left\_node\_lower\_bound} and \emph{right\_node\_lower\_bound}, instead of a general node lower bound, because during experimentation, we obtained two different lower bounds among the child nodes.

The actions are the node selection decisions at a node.  As opposed to unconstrained node selectors---which can choose any open node s- we constrain our node selector's action space $a_t$ to the selection of direct child nodes.  This leads to the restricted action space $\{L, R, B\}$, where $L$ is the left child, $R$ is the right child and $B$ are both children.

In order to train a policy by imitation learning, we require training data from the expert.  We obtain this training data, in the form of sampled state-action pairs, by running a MIP solver (specifically SCIP), and recording the features from the search process and the node selection decisions at each sampled node.  We now explain the process.

\subsection{Data Collection and Processing}
	Our sampling process is similar to the prior work of \mbox{He et al. \citep{He2014:learning}}, with two major differences.
The first is that our policy learns \emph{only} to choose which of a node's children it should select; it does not consider the sub-tree below either child.  This encourages finding solutions quickly, as opposed to attempting to learn a breadth-first search-like method.

The second difference from previous work is that we generalise the node selection process by sampling paths that lead to the best $k$ solutions, instead of only the top solution.  The reason for this is to obtain different state-action pairs that lead to good solutions compared to only using the top solution, in order to aid learning within a deep learning context.
In more detail, the best $k$ solutions are chosen in the same order as output from SCIP; we therefore adopt SCIP's solution ranking which is based on the optimality gap.
Thus, whereas \mbox{He et al. \citep{He2014:learning}} check whether the current node is in the path to the best solution, we check whether the left and right children of the current node are in a path that leads to one of the best $k$ solutions.  If that is the case, then we associate the label of the current node as `B'; if not, we check the left child or right child and associate the appropriate label (`L' or `R' respectively).  If neither are in such a path, then the node is not sampled.

Pre-processing of the dataset is done by removing features that do not change and standardising every non-categorical feature.  We then feed it to the imitation learning component, described next, after the example.

\end{paracol}
\nointerlineskip
\begin{specialtable}[H]
\setlength{\tabcolsep}{0.8mm} 
\widetable
\caption{Features that define a state. Variable features from  Gasse et al. \citep{DBLP:conf/nips/GasseCFC019}.}
\label{tab:features}
\begin{tabular}{lll}
\toprule
\textbf{Category}                   & \textbf{Feature}                & \textbf{Description}                                                     \\ \midrule
\multirow{13}{*}{Variable features} & type                            & Type (binary, integer, impl. integer, continuous) as a one-hot encoding \\ \cmidrule(l){2-3} 
                                    & coef                            & Objective coefficient, normalized                                       \\ \cmidrule(l){2-3} 
                                    & has\_lb                         & Lower bound indicator                                                   \\ \cmidrule(l){2-3} 
                                    & has\_ub                         & Upper bound indicator                                                   \\ \cmidrule(l){2-3} 
                                    & sol\_is\_at\_lb                 & Solution value equals lower bound                                       \\ \cmidrule(l){2-3} 
                                    & sol\_is\_at\_ub                 & Solution value equals upper bound                                       \\ \cmidrule(l){2-3} 
                                    & sol\_frac                       & Solution value fractionality                                             \\ \cmidrule(l){2-3} 
                                    & basis\_status                   & Simplex basis status (lower, basic, upper, zero) as a one-hot encoding  \\ \cmidrule(l){2-3} 
                                    & reduced\_cost                   & Reduced cost, normalized                                                \\ \cmidrule(l){2-3} 
                                    & age                             & LP age, normalized                                                      \\ \cmidrule(l){2-3} 
                                    & sol\_val                        & Solution value                                                          \\ \cmidrule(l){2-3} 
                                    & inc\_val                        & Value in incumbent                                                      \\ \cmidrule(l){2-3} 
                                    & avg\_inc\_val                   & Average value in incumbents                                             \\ \midrule
\multirow{8}{*}{Node features}      & left\_node\_lb                  & Lower (dual) bound of left subtree                                      \\ \cmidrule(l){2-3} 
                                    & left\_node\_estimate            & Estimate solution value of left subtree                                 \\ \cmidrule(l){2-3} 
                                    & left\_node\_branch\_bound       & Branch bound of left subtree                                            \\ \cmidrule(l){2-3} 
                                    & left\_node\_is\_prio            & Branch rule priority indication of left subtree                         \\ \cmidrule(l){2-3} 
                                    & right\_node\_lb                 & Lower (dual) bound of right subtree                                     \\ \cmidrule(l){2-3} 
                                    & right\_node\_estimate           & Estimate solution value of right subtree                                \\ \cmidrule(l){2-3} 
                                    & right\_node\_branch\_bound      & Branch bound of right subtree                                           \\ \cmidrule(l){2-3} 
                                    & right\_node\_is\_prio           & Branch rule priority indication of right subtree                        \\ \midrule
\multirow{8}{*}{Global features}    & global\_upper\_bound            & Best feasible solution value found so far                               \\ \cmidrule(l){2-3} 
                                    & global\_lower\_bound            & Best relaxed solution value found so far                                \\ \cmidrule(l){2-3} 
                                    & integrality\_gap                & Current integrality gap                                                 \\ \cmidrule(l){2-3} 
                                    & gap\_is\_infinite               & Gap is infinite indicator                                               \\ \cmidrule(l){2-3} 
                                    & depth                           & Current depth                                                           \\ \cmidrule(l){2-3} 
                                    & n\_strongbranch\_lp\_iterations & Total number of simplex iterations used so far in strong branching      \\ \cmidrule(l){2-3} 
                                    & n\_node\_lp\_iterations         & Total number of simplex iterations used so far for node relaxations     \\ \cmidrule(l){2-3} 
                                    & max\_depth                      & Current maximum depth                                                   \\ \bottomrule
\end{tabular}

\end{specialtable}
\begin{paracol}{2}
\switchcolumn

\vspace{-6pt}

\subsection{{Example of Data Collection}}

{We here give an example of the data collection and processing on a toy problem with two variables.  Let $k = 3$ and suppose the known best 3 solutions are: $(x_1, x_2) \in { (1, 2), (0, 0), (4, 3) }$.  The sampling process begins by activating the solver.}

\begin{enumerate}
{
    \item The solver solves the relaxation of current (root) node and variable selection tells it to branch on variable $x_1$ with bounds $x_1 \leq 2$ and $x_1 \geq 3$.  In this case we know that both children are in the path of best 3 solutions, so the label here is `B'. Note that we just found a state-action pair and so we save it.
    \item The solver continues its process as usual and decides to enter the left child.
    \item Again, the solver solves the relaxation and applies variable selection, the following two child nodes are generated with bounds: $x_2 \leq 3$ and $x_2 \geq 4$.  This case is interesting, because only the left child can lead to best 3 solutions, so the label here is 'L'. Again we save this state-action pair.
    \item Since we do not assume control of the solver in the sampling process, let us suppose the solver enters the right child (recall the best 3 solutions are not here, but the solver does not know that).
    \item Two additional child nodes are generated: $x_1 \leq 0$ and $x_1 \geq 1$.  The left node has the following branched variable conditions in total: $x_1 \leq 2$, $x_2 \geq 4$ and $x_1 \leq 0$. and right node has: $x_1 \leq 2$, $x_2 \geq 4$ and $x_1 \geq 1$.  None of these resolve to any of the best 3 solutions.  Thus we do not sample anything here.
    \item The solver continues to select the next node.
    \item We continue to proceed with the above until the solver stops. Now we have sampled state-action pair from one training instance.  This process is repeated for many \mbox{training instances}.}
\end{enumerate}

\subsection{Machine Learning Model}
Our machine learning model is a standard fully-connected multi-layer perceptron
with $H$ hidden layers, $U$ hidden units per layer, ReLU activation layers and Batch Normalisation layers after each activation layer, following a Dropout layer with dropout rate ${p_d}$.  Figure~\ref{fig:ml_arch} gives a visual overview of the operations within a hidden layer.

\begin{figure}[H]
    \includegraphics[width=\linewidth]{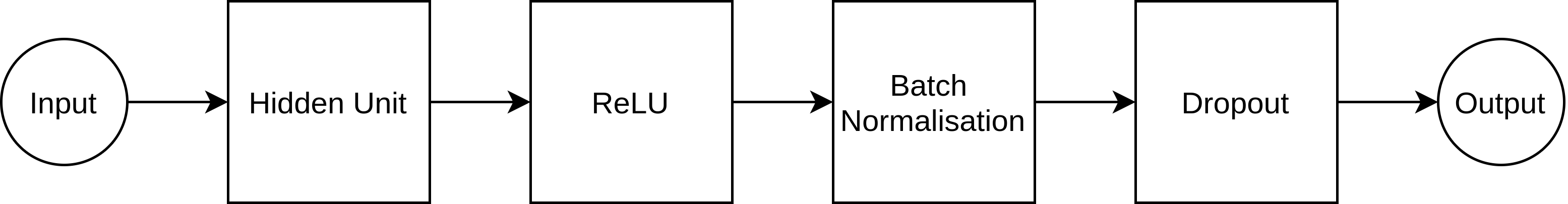}
    \caption{Operations within a hidden layer of the network.}
    \label{fig:ml_arch}
\end{figure}

We obtain the model architecture parameters and learning rate $\rho$ using the hyperparameter optimisation algorithm
\citep{bergstra2013making}.  Since during pre-processing features that have constant values are removed, the number of input units can change across different problems.  For example, in a fully binary problem, the features \emph{left\_node\_branch\_bound} and \emph{right\_node\_branch\_bound} are constants ($0$ and $1$ respectively), while for a general mixed-integer problem this is not the case.  The number of output units is three.  The cross-entropy loss is optimised during training with the Adam algorithm \citep{DBLP:journals/corr/KingmaB14}.

During policy evaluation, the action $B$ (`both') can result in different operations, as seen in Table~\ref{table:param_settings}.  We define \heur{PrioChild}, \heur{Second} and \heur{Random} as possible operations.  \heur{PrioChild} selects the priority child as indicated by the variable selection heuristic (\ie the brancher);
\heur{Second} selects the next best scoring action from the ML policy; \heur{Random} selects a random child.  Additionally, when the solver is at a leaf and there is no child to select, then we define three more operations.  These are \heur{RestartDFS}, \heur{BestEstimate} and \heur{Score}.  The first two are baseline node selectors from SCIP \citep{scip6};
\heur{Score} selects the node which obtained the highest score so far as calculated by our node selection policy.
 
\begin{specialtable}[h]
\setlength{\tabcolsep}{15.6mm} 
    \caption{Parameter settings for our node selection and pruning policy.  Policies are denoted by \policy{ML\_} followed by up to three letters: \policy{ML\_\{on\_both\}\{on\_leaf\}\{prune\_on\_both\}}.
 {Table~\protect\ref{table:learning-configurations} enumerates the resulting policies.}}
\label{table:param_settings}
\begin{tabular}{ll}
\toprule
\textbf{Parameter} & \textbf{Domain}                     \\ \midrule
on\_both           & \{PrioChild, Second, Random\}       \\
on\_leaf           & \{RestartDFS, BestEstimate, Score\} \\
prune\_on\_both    & \{True, False\}                     \\ \bottomrule
\end{tabular}
\end{specialtable}

Obtaining the node pruning policy is similar to obtaining the node selection policy.  The difference is that the node pruning policy also prunes the child that is ultimately not selected by the node selection policy.  If \emph{prune\_on\_both} = True, then this results in diving only once and then terminating the search.  Otherwise, the nodes initially not selected after the action $B$ are still explored.  The resulting solving process is thus approximate, since we cannot guarantee that the optimal solution is not pruned.

To summarise, as seen in Figure~\ref{fig:flowchart}, we use our learned policy in two ways: the first is heuristic by committing to pruning of nodes, whereas the second is exact: when reaching a leaf, we backtrack up the branch-and-bound tree to use a different strategy.  Source code is available at: \url{www.doi.org/10.4121/14054330}; note that due to licensing restrictions, SCIP must be obtained separately from \url{www.scipopt.org} (accessed 01-Sep-21).

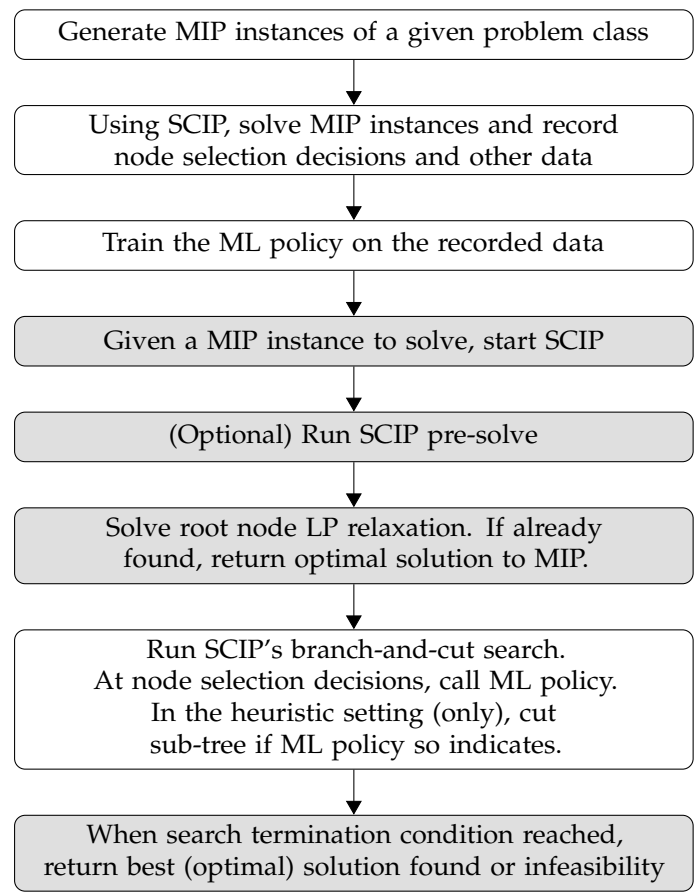
\begin{figure}[H]
\colorlet{lcfree}{Green3}
\colorlet{lcnorm}{black}
\colorlet{lccong}{Red3}
\begin{tikzpicture}[%
    >=triangle 60,
    start chain=going below,
    node distance=6mm and 60mm,
    every join/.style={norm},
    ]
\tikzset{
  base/.style={draw, on chain, on grid, align=center, minimum height=4ex},
  proc/.style={base, rectangle, text width=25em},
  test/.style={base, diamond, aspect=2, text width=5em},
  term/.style={proc, rounded corners},
  grey/.style={proc, rounded corners, fill=gray!25},
  coord/.style={coordinate, on chain, on grid, node distance=6mm and 25mm},
  nmark/.style={draw, cyan, circle, font={\sffamily\bfseries}},
  norm/.style={->, draw, lcnorm},
  free/.style={->, draw, lcfree},
  cong/.style={->, draw, lccong},
  it/.style={font={\small\itshape}}
}
\node [term]            {Generate MIP instances of a given problem class};
\node [term, join]      {Using SCIP, solve MIP instances and record node selection decisions and other data};
\node [term, join]      {Train the ML policy on the recorded data};
\node [grey, join]      {Given a MIP instance to solve, start SCIP};
\node [grey, join]      {(Optional) Run SCIP pre-solve};
\node [grey, join]      {Solve root node LP relaxation.  If already found, return optimal solution to MIP.};
\node [term, join]      {Run SCIP's branch-and-cut search. \\ At node selection decisions, call ML policy. \\ In the heuristic setting (only), cut sub-tree if ML policy so indicates.};
\node [grey, join]      {When search termination condition reached, \\ return best (optimal) solution found or infeasibility};
\end{tikzpicture}
    \caption{Flowchart of the ML-based approximate node pruning, for a given problem class.  \mbox{Shaded boxes} are standard SCIP steps.  The first three steps are training, which is performed once for a problem class; the remaining steps are the online solving, which is performed for a given input MIP instance to be solved.}
    \label{fig:flowchart}
\end{figure}

\section{Empirical Results}
\label{sec:results}
This section studies empirically the node selection policies described in Section~\ref{sec:method}.  \mbox{The goal} is to explore the effectiveness of the learning and of the learned policy within a \mbox{MIP solver}.

The following standard NP-hard problem classes were tested: set cover, maximum independent set, capacitated facility location and combinatorial auctions.  The instances were generated by the generator of  \mbox{Gasse et al. \citep{DBLP:conf/nips/GasseCFC019}} with default settings.  The classes differ from each other in terms of constraints structure, existence of continuous variables, existence of non-binary integer variables, and direction of optimisation.

For the MIP branch-and-bound framework we use SCIP version 6.0.2. {(Note that SCIP version 7, released after we commenced this work, does not bring any major improvements to its MIP solving.)}  As noted earlier, SCIP is an open source MIP solver, allowing us access to its search process.  Further, SCIP is regarded as the most sophisticated and fastest such MIP solver.
The machine learning model is implemented in PyTorch
\citep{DBLP:conf/nips/PaszkeGMLBCKLGA19}, and interfaces with SCIP's C code via PySCIPOpt 2.1.6.

For every problem, we show the learning results, \ie how well the policy is learned, and the MIP benchmarking results, \ie how well the MIP solver does with the learned policy.
We compare the policy evaluation results with various node selectors in SCIP, namely \heur{BestEstimate} with plunging (the SCIP default), \heur{RestartDFS} and \heur{DFS}.
Additionally, we compare our results with the node selector and pruner from He et al. \citep{He2014:learning}, with both the original SCIP 3.0 implementation by those authors (\heur{He}) and with the a re-implementation in SCIP 6 developed by us (\heur{He6}).  He et al. \citep{He2014:learning} has three policies: selection only (S), pruning only (P) and both (B).  For exact solutions, we only use (S). For the first solution found at a leaf, we compare (S) and (B). For the experiments with a time limit, we compare (P) and (B).
Table~\ref{tab:methods-summary} summarises the node selectors compared.

\begin{specialtable}[H]    
\setlength{\tabcolsep}{9.5mm} 

\caption{Summary of methods compared in the experiments.}
\label{tab:methods-summary}
\begin{tabular}{lllllll}
\toprule
\textbf{Method} & \textbf{Origin} & \textbf{Exact?} \\ \midrule
\heur{BestEstimate} & SCIP \citep{scip6} & yes \\
\heur{DFS}          & SCIP \citep{scip6} & yes \\
\heur{RestartDFS}   & SCIP \citep{scip6} & yes \\
\heur{He} (select only)  & He et al. \citep{He2014:learning} original       & yes \\
\heur{He} (prune only)  & He et al. \citep{He2014:learning} original       & no \\
\heur{He} (both)  & He et al. \citep{He2014:learning} original       & no \\
\heur{He6} (select only) & He et al. \citep{He2014:learning} re-implemented & yes \\
\heur{He6} (prune only) & He et al. \citep{He2014:learning} re-implemented & no \\
\heur{He6} (both) & He et al. \citep{He2014:learning} re-implemented & no \\
\policy{ML\_$\cdot\cdot\cdot$} heuristic & this article & no \\
\policy{ML\_$\cdot\cdot\cdot$} exact     & this article & yes \\
\bottomrule
\end{tabular}
\end{specialtable}

We train on 200 training instances, 35 validation instances and 35 testing instances across all problems.  These provide sufficient state-action pairs to power the machine learning model.  The number of obtained samples (state-action pairs) differs per problem.  {The process of obtaining samples from each training instance was described in \mbox{Section~\ref{sec:method}}.}  For every problem, we use the $k = 10$ best solutions to gather the state-action pairs; \mbox{for maximum} independent set we also experiment with $k = 40$.  
The higher the value of $k$, the more data we can collect and the better one can generalise the ML model; we discuss further in Section~\ref{sec:discuss}.  We selected $k=10$ having tried lower values ($k=1, 2, 5$) and found inferior results in initial experiments.  Due to computational limitations on training time, higher values of $k$ were not feasible across the board.

Based on initial trials and the hyperparameter optimisation, we use a batch size of 1024, dynamically lower the learning rate after 30 epochs and terminate training after another 30 epochs if no improvement was found. During training, the validation loss is optimised.  The maximum number of epochs is 200.


We evaluate a number of different settings for our node selection and pruning policy, as seen in Table~\ref{table:param_settings}.  This leads to nine different configurations for the node selection policy and twelve different configurations for the node pruning policy.  Note that for the node pruning policy, when \textit{prune\_on\_both} is true, then optimisation terminates when a leaf is found; thus the parameter value for \textit{on\_leaf} does not matter.  We refer to our policies as \policy{ML\_\{on\_both\}\{on\_leaf\}\{prune\_on\_both\}}.
For example, \policy{ML\_PB} denotes the node pruning policy that uses \emph{PrioChild} for \param{on\_both} and \emph{BestEstimate} for \param{on\_leaf}.  {Table~\ref{table:learning-configurations} enumerates the set of configurations for the learned policies.}

\begin{specialtable}[H]
\caption{{The node selection (top) and pruning (bottom) policy configurations in full.}}
\label{table:learning-configurations}
\setlength{\tabcolsep}{9mm} 
\begin{tabular}{@{}lllllll@{}}
\toprule
\textbf{{Method}} & \textbf{{on\_both}} & \textbf{{on\_leaf}} & \textbf{ {prune\_on\_both}} \\ \midrule
\policy{ML\_PR} & PrioChild & RestartDFS   & false \\
\policy{ML\_SR} & Second    & RestartDFS   & false \\
\policy{ML\_RR} & Random    & RestartDFS   & false \\
\policy{ML\_PB} & PrioChild & BestEstimate & false \\
\policy{ML\_SB} & Second    & BestEstimate & false \\
\policy{ML\_RB} & Random    & BestEstimate & false \\
\policy{ML\_PS} & PrioChild & Score        & false \\
\policy{ML\_SS} & Second    & Score        & false \\
\policy{ML\_RS} & Random    & Score        & false \\
\midrule
\policy{ML\_PRF} & PrioChild & RestartDFS   & false \\
\policy{ML\_SRF} & Second    & RestartDFS   & false \\
\policy{ML\_RRF} & Random    & RestartDFS   & false \\
\policy{ML\_PBF} & PrioChild & BestEstimate & false \\
\policy{ML\_SBF} & Second    & BestEstimate & false \\
\policy{ML\_RBF} & Random    & BestEstimate & false \\
\policy{ML\_PSF} & PrioChild & Score        & false \\
\policy{ML\_SSF} & Second    & Score        & false \\
\policy{ML\_RSF} & Random    & Score        & false \\
\policy{ML\_P$\cdot$T}& PrioChild &--  & true  \\
\policy{ML\_S$\cdot$T}& Second    &--  & true  \\
\policy{ML\_R$\cdot$T}& Random    &--  & true  \\
\bottomrule
\end{tabular}

\end{specialtable}

\textls[-40]{In the following we report three sets of experiments, and then a subsequent \mbox{fourth experiment}:}
\begin{enumerate}
    \item \textbf{First solution.} In the first experiment, we examine the solution quality in terms of the optimality gap and the solving time of the first solution found at a leaf node.  Recall that the optimality gap is the difference between the objective function value of the best found solution and that of the optimal solution.  Note that it is possible an infeasible leaf node is found, in that case, a solution is returned that was found prior to the branch-and-bound process, through heuristics in-built in SCIP.
    
    \item \textbf{Optimal solution.} In the second experiment, we evaluate the policy on every problem in terms of the arithmetic mean solving time of each node selector.  That is, the total time to find an optimal solution and prove its optimality.
    
    \item \textbf{Limited time.} In the third experiment, we select one ML policy, based on the (lowest) harmonic mean between the solving time and optimality gap. For each instance, \mbox{we run} the solver on each baseline with a time limit equal to the solving time of the selected ML policy and present the obtained optimality gaps.  We also report the initial optimality gap obtained by the solver before branch-and-bound is applied.

    \item \textbf{Imitation.} In the fourth experiment, we analyze the behaviour of the learned policy during the first plunge, and compare it to SCIP's default \heur{BestEstimate} rule.  We do this in detail on one set of instances.
\end{enumerate}

By \termdefn{solving time} we mean the gross difference in time between starting SCIP and terminating it.  When we terminate SCIP depends on the task of the experiment:
\begin{itemize}
    \item Solve to optimality and record solving time: termination condition is finding an optimal solution.
    \item Solve until we find a solution in a leaf node (or by SCIP's built-in heuristics if the leaf node reached is infeasible) and record the solving time and optimality gap: termination condition is finding a first leaf node.
    \item Solve until a certain time limit and record optimality gap: termination condition is the time limit itself.
\end{itemize}

We
apply the policies on instances of two different difficulties.

    First, easy instances, which can be solved within 15 min.
    Second,
    hard instances, where we set a solving time limit to one hour.  Here, for all
    experiments we substitute the integrality gap for the optimality gap, because the optimal solution is not known for every hard instance.  An alternative approach is to compare directly the primal solution quality (see Section~\ref{sec:discuss}).
    Additionally with hard instances, for Experiment~1, instead of checking the solving time, we check the integrality gap.

\textls[-30]{Table~\ref{table:ml_overview} provides an overview of the machine learning parameters and results.  \mbox{The baseline}} accuracy (column 2) is what the accuracy would have been if each sample is classified as the majority class.  The test accuracy (column 3) is the classification accuracy on the test dataset.  Note that $k=40$ is included in the maximum independent set instances: see Appendix~\ref{section:indset}.
The best performing ML model is the model with the settings that achieve the lowest validation loss.

\begin{specialtable}[H]
\setlength{\tabcolsep}{3.6mm} 
       \caption{ML parameters and prediction results.
The baseline accuracy is predicting everything as the majority class.}
\label{table:ml_overview}
\begin{tabular}{lllllll}
\toprule
\textbf{Problem}                    & \textbf{Base} & \textbf{Test} & \boldmath{$H$} & \boldmath{$U$} & \boldmath{$p_d$} & \boldmath{$\rho$} \\ \midrule
Set cover                                        & 0.575                      & 0.764                     & 1          & 49         & 0.445         & 0.253                        \\ \midrule
Maximum independent set (10)                                & 0.923                      & 0.922                     & 1          & 25         & 0.266         & 0.003                        \\
Maximum independent set (40)                                              & 0.895                      & 0.899                     & 1          & 42         & 0.291         & 0.003                        \\ \midrule
Capacitated facility location                    & 0.731                      & 0.901                     & 3          & 20         & 0.247         & 0.008                        \\ \midrule
Combinatorial auctions                           & 0.570                      & 0.717                     & 1          & 9          & 0.169         & 0.002                        \\ \bottomrule
\end{tabular}
\end{specialtable}

The experiments reported in Sections~\ref{sec:results:setccover}--\ref{sec:results:setcoverhard} are run on a machine with an Intel i7 8770K CPU at 3.7--4.7 GHz, NVIDIA RTX 2080 Ti GPU and 32GB RAM.  For the hard instances, the default SCIP solver settings are used.  For the other instances, pre-solving and primal heuristics are turned off, to better capture the effect of the node selection policy. 
We use the shifted geometric mean (shift of $1$) as the average across all metrics.  This is standard practice for MIP benchmarks \citep{achterberg2007constraint}.

\subsection{{Summary of Results}}
\label{sec:results:summary}

{The experiments can be summarised as follows:}
\begin{itemize}
    \item {\textbf{Set cover:} \heur{BestEstimate} has the lowest mean solving time.  \policy{ML\_RB} is second, but is not statistically significantly slower.  The policies of He et al. \citep{He2014:learning} are slow.  When a time limit has imposed, \policy{ML\_SRF} has the lowest harmonic mean of solving time and optimality gap; this is statistically significant compared to all others.}
    \item  {\textbf{Other instances:} On maximum independent set, \heur{DFS} dominates and the ML policies are relatively poor.  Only on this dataset are our policies slower (slightly) than \heur{He6}; in all other cases they dominate.  On capacity facility location and on combinatorial auctions, the ML policies and \heur{BestEstimate} respectively are fastest, but not statistically significantly so.}
    \item {\textbf{Hard set cover:} \heur{BestEstimate} has the lowest mean solving time, but no ML policy was statistically significantly slower.  When a time limit is imposed, \policy{ML\_PST} has the lowest harmonic mean of solving time and optimality gap; this is statistically significant compared to all others.}
    \item  {\textbf{Imitation quality:} When the policies are analysed in detail, the \heur{PrioChild} policies \policy{ML\_P$\cdot\cdot$} are found to accurately imitate \heur{BestEstimate}.}
\end{itemize}

\subsection{Set Cover Instances}
\label{sec:results:setccover}
These instances consist of 2000 variables and 1000 constraints forming a pure binary minimisation problem. We sampled 17,254 state-action pairs on the training instances, 2991 on the validation instances and 3218 on the test instances. The model achieves a testing accuracy of 76.4\%, with a baseline accuracy of 57.5\%.

Figure~\ref{fig:setcover_pareto} shows the mean solving time against the mean optimality gap of the first solution obtained by the baselines and ML policies at a leaf node.  Note that the only parameter
that is influential for the ML solver is the first parameter, \ie \textit{on\_both}. {(\mbox{The second} parameter \textit{on\_leaf} and the third
\textit{prune\_on\_both} do not influence the solving time or quality of the first solution, since the search terminates at the first found leaf that is found.)}  The policy of He et al. \citep{He2014:learning} is not included here due to its substantial outliers.  We see here that our ML policy obtains a lower optimality gap at the price of a higher solving time for the first solution.

\begin{figure}[H]
        \includegraphics[width=0.95\linewidth,clip,trim={0 0cm 0 1.2cm}]{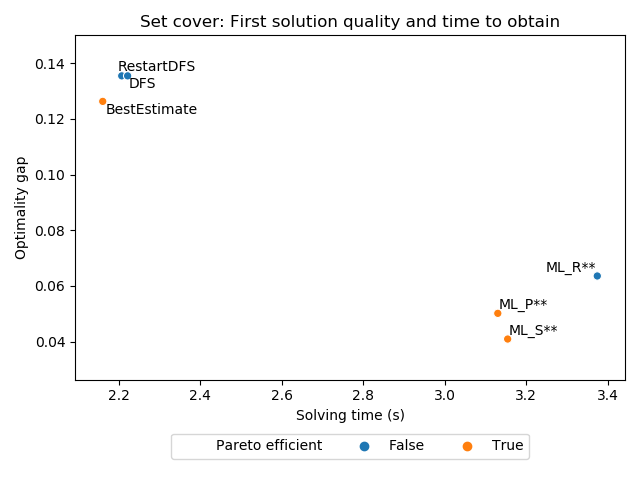}
        \caption{Set cover: mean solving time vs.\@ mean optimality gap of the first solution found at a leaf node.
        Orange points are Pareto-efficient, blue are not.}
        \label{fig:setcover_pareto}
\end{figure}

Table~\ref{table:setcover_selection} reports the mean solving time and explored nodes of various exact node selection strategies.  \heur{BestEstimate} achieves the lowest mean solving time at $26.4$ s; \policy{ML\_RB} comes next at $35.7$ s.  The policies of He et al. \citep{He2014:learning}--both the original and our re-implementation--are markedly slower than all other methods.  

We conducted a pair-wise t-test between the mean solving time of \heur{BestEstimate} and the mean solving time of the other policies.  We can only reject the null hypothesis of equal means with \emph{p}-value below 0.1 for \policy{ML\_RR} (\emph{p}-value: 0.08). For the rest of our ML policies, we cannot reject the null hypothesis of equal means.
We also conducted a pair-wise t-test between the mean number of explored nodes of \heur{BestEstimate} and the mean number of explored nodes of the other policies. Again, we cannot reject the null hypothesis of equal means with \emph{p}-value below 0.05 for any of our ML policies (lowest observed \emph{p}-value: 0.34).

Table~\ref{table:policy_eval_setcover} reports
the mean optimality gap of the baselines under a \emph{time limit} for each instance.  The time limit for each instance is based on the solving time of the ML policy that achieved the lowest harmonic mean between the mean solving time and mean optimality gap across all instances.  This time limit fosters comparison because it is certain to be neither excessively low, so that the different methods would accomplish little, nor excessively high, so that all methods would find an optimal solution.
In the case of the set cover instances,
\policy{ML\_SRF} has the lowest harmonic mean and also achieves the lowest mean optimality gap. 
We conducted a pairwise t-test between the mean optimality gap of our best ML policy and the mean optimality gap of each baseline.  We can reject the null hypothesis of equal means with \emph{p}-value below 0.005 for all baselines.
This shows that \policy{ML\_SRF}'s smaller optimality gap is significant.

The initial optimality gap obtained by the solver before branch-and-bound is $2.746$.
This shows that applying branch-and-bound--with whatever node selection strategy--to find a solution has a significant difference.

\begin{specialtable}[H]
\setlength{\tabcolsep}{11mm} 
\caption{Set cover instances: mean solving time and explored nodes for various node selection strategies. Pair-wise t-tests compared to \heur{BestEstimate}: `***' $p < 0.001$, `**' $p < 0.01$, `*' $p < 0.05$, `$\cdot$' $p < 0.1$.}
\label{table:setcover_selection}
\begin{tabular}{lll}\toprule
\textbf{Strategy} &\textbf{Solving Time (s)} &\textbf{Explored Nodes} \\\midrule
BestEstimate &\textbf{26.39} &\textbf{4291} \\
DFS &47.10 * &9018 * \\
He (S)  &499.00 *** &47116 *** \\
He6 (S) &85.73 ** &18386 ** \\
\policy{ML\_PB} &36.82 &4573 \\
\policy{ML\_PR} &38.82 &5241 \\
\policy{ML\_PS} &39.73 &5295 \\
\policy{ML\_RB} &35.68 &4663 \\
\policy{ML\_RR} &40.09 $\cdot$ &5666 \\
\policy{ML\_RS} &37.13 &4799 \\
\policy{ML\_SB} &35.90 &4408 \\
\policy{ML\_SR} &36.64 &4776 \\
\policy{ML\_SS} &37.61 &4923 \\
RestartDFS &45.30 * &8420 * \\
\bottomrule
\end{tabular}

\end{specialtable}

\vspace{-6pt}

\begin{specialtable}[H]    
\setlength{\tabcolsep}{23mm} 
\caption{Set cover: ML model with \textit{on\_both} = Second, \textit{on\_leaf} = RestartDFS and \textit{prune\_on\_both} = False against baselines, with equal time limits for each problem.
Pairwise t-tests against the ML policy: `***' $p < 0.001$, `**' $p < 0.01$, `*' $p < 0.05$, `$\cdot$' $p < 0.1$.}
\label{table:policy_eval_setcover}
\begin{tabular}{ll}\toprule
\textbf{Strategy} &\textbf{Optimality Gap} \\\midrule
BestEstimate &0.1767 ** \\
DFS     &0.0718 *** \\
He6 (P) &0.7988 *** \\
He6 (B) &1.1040 *** \\
\policy{ML\_SRF} &\textbf{0.0278} \\
RestartDFS &0.0741 *** \\
\bottomrule
\end{tabular}
\end{specialtable}

\subsection{Three Further Problem Classes}
\label{sec:results:more}

The
appendix gives detailed results for the maximum independent set, capacitated facility location, and combinatorial auctions problems.
Briefly, on maximum independent set, baseline \heur{DFS} dominates, and the ML policies are relatively poor; this is the only problem class where \heur{He6} outperforms our policies slightly (although we outperform the original \heur{He} easily).  On capacitated facility location, our ML policies give the best results on solving time and on explored nodes, although the results are not significantly better than \heur{BestEstimate} according to the t-test.  On combinatorial auctions, \heur{BestEstimate} dominates, although its explored nodes are not significantly fewer than our ML policies.

When a time limit is applied, our best ML policy either achieves the best optimality gap (as with set cover problems), or is Pareto-equivalent to the baseline which does (interestingly, this is \heur{DFS}, never \heur{BestEstimate} in these problems).  This in in contrast to \emph{He6}, which in all problems is significantly poorer in optimality gap obtained.

\subsection{Set Cover: Hard Instances}
\label{sec:results:setcoverhard}

To assess how the ML policies perform on hard instances, we use the same trained model of the ML policies that were previously trained on the easier set cover instances.  \mbox{The hard} instances consist of 4000 variables and 2000 constraints, while the easier set cover instances had 2000 variables and 1000 constraints.  We evaluated 10 hard instances (due to computational limitations) and focused on \heur{BestEstimate} as a baseline on the node selection policy.
Primal heuristics  {and pre-solving} were enabled.

For the pruning policies, Figure~\ref{fig:setcover_hard_pareto} shows the solving time plotted against the integrality gap of the first solution obtained by the baselines and ML policies.
As before, we see the same trend where the ML policies find a lower gap at the cost of a higher solving time.
See Table~\ref{table:policy_eval_setcover_hard} and Figure~\ref{fig:setcover_hard_approx_boxplot} for the integrality gap of the baselines using a time limit for each instance.  In this case, \policy{ML\_PST} has the lowest harmonic mean between the mean solving time and mean integrality gap of all ML policies.  \policy{ML\_PST} achieves a significantly lower integrality gap compared to the baselines: we conducted a pairwise t-test between the mean optimality gap of our best ML policy and the mean optimality gap of each baseline.  We can reject the null hypothesis of equal means with \emph{p}-value below 0.001 for all baselines.

For the node selection policies, we set the time limit to one hour per problem.  \mbox{Figure~\ref{fig:setcover_hard_solved_instances}} shows the number of solved instances per policy, and Figure~\ref{fig:setcover_hard_optimality_boxplots} shows boxplots of the integrality gaps for each policy.  We use integrality gap here, because we do not know the optimal objective value for all instances.
Table~\ref{table:setcover_hard_selection} shows the mean solving time, explored nodes and integrality gap of various node selection strategies.  \heur{BestEstimate} achieves the lowest mean solving time at $2256.7$ seconds and integrality gap at $0.0481$.  \policy{ML\_SB} has the lowest number of explored nodes at $109298$.
A caveat over the reduced number of nodes of the best ML policies versus \heur{BestEstimate} comes from the solver timing out on many instances--the instances being hard.  Fewer nodes could be due to the overhead that the ML classifier requires at each node.

We conducted a pair-wise t-test between the mean solving time of \heur{BestEstimate} and the mean solving time of the other policies.  We cannot reject the null hypothesis of equal means with \emph{p}-value below 0.1 for all our ML policies (lowest observed \emph{p}-value: 0.64).
We also conducted a pair-wise t-test between the mean number of explored nodes of \heur{BestEstimate} and the mean number of explored nodes of the other policies. We can reject the null hypothesis of equal means with \emph{p}-value below 0.05 for \policy{ML\_PB}, \policy{ML\_RB} and \policy{ML\_SB}.
Lastly, we conducted a pair-wise t-test between the mean integrality gap of \heur{BestEstimate} and the mean integrality gap of the other policies. We can reject the null hypothesis of equal means with \emph{p}-value below 0.1 for \policy{ML\_RS} and \policy{ML\_SS}.
Summarising, \policy{ML\_SB} has a statistically significant fewer number of nodes, while not having a statistically significant higher time or integrality gap than \heur{BestEstimate}.

Unlike the easier set cover instances, in all the hard instances, the solver could only obtain an integrality gap of infinity on the first feasible solution.  Hence
we cannot compare the initial integrality gap to the found integrality gaps during branch and bound.
A possible reason for the solver's integrality gap of infinity could be that the first solution is found before the root LP relaxation was solved.

\begin{figure}[H]
        \includegraphics[width=0.95\linewidth,clip,trim={0 0cm 0 1.2cm}]{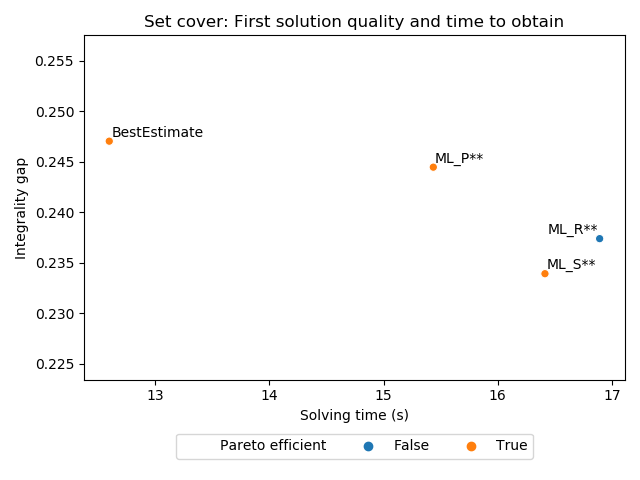}
        \caption{Hard set cover: mean solving time vs.\@ mean integrality gap of the first solution found at a leaf node.
        Orange points are Pareto-efficient, blue are not.}
        \label{fig:setcover_hard_pareto}
\end{figure}

\vspace{-6pt}
    
\begin{specialtable}[H]
\setlength{\tabcolsep}{23.4mm} 
\caption{Hard set cover: ML model with \textit{on\_both} = PrioChild, \textit{on\_leaf} = Score and \textit{prune\_on\_both} = True against the baselines, with equal time limits for each problem.  Pairwise t-tests against the ML policy: `***' $p < 0.001$, `**' $p < 0.01$, `*' $p < 0.05$, `$\cdot$' $p < 0.1$.}
\label{table:policy_eval_setcover_hard}
\begin{tabular}{ll}\toprule
\textbf{Strategy} &\textbf{Integrality Gap} \\\midrule
BestEstimate &1.3678 *** \\
DFS     &0.3462 *** \\
He6 (B) &1.8315 *** \\
He6 (P) &1.6835 *** \\
\policy{ML\_PST} &\textbf{0.2445} \\
RestartDFS &0.3489 *** \\
\bottomrule
\end{tabular}
\end{specialtable}
\vspace{-6pt}

\begin{figure}[H]
    \includegraphics[width=1.025\linewidth,
    clip]{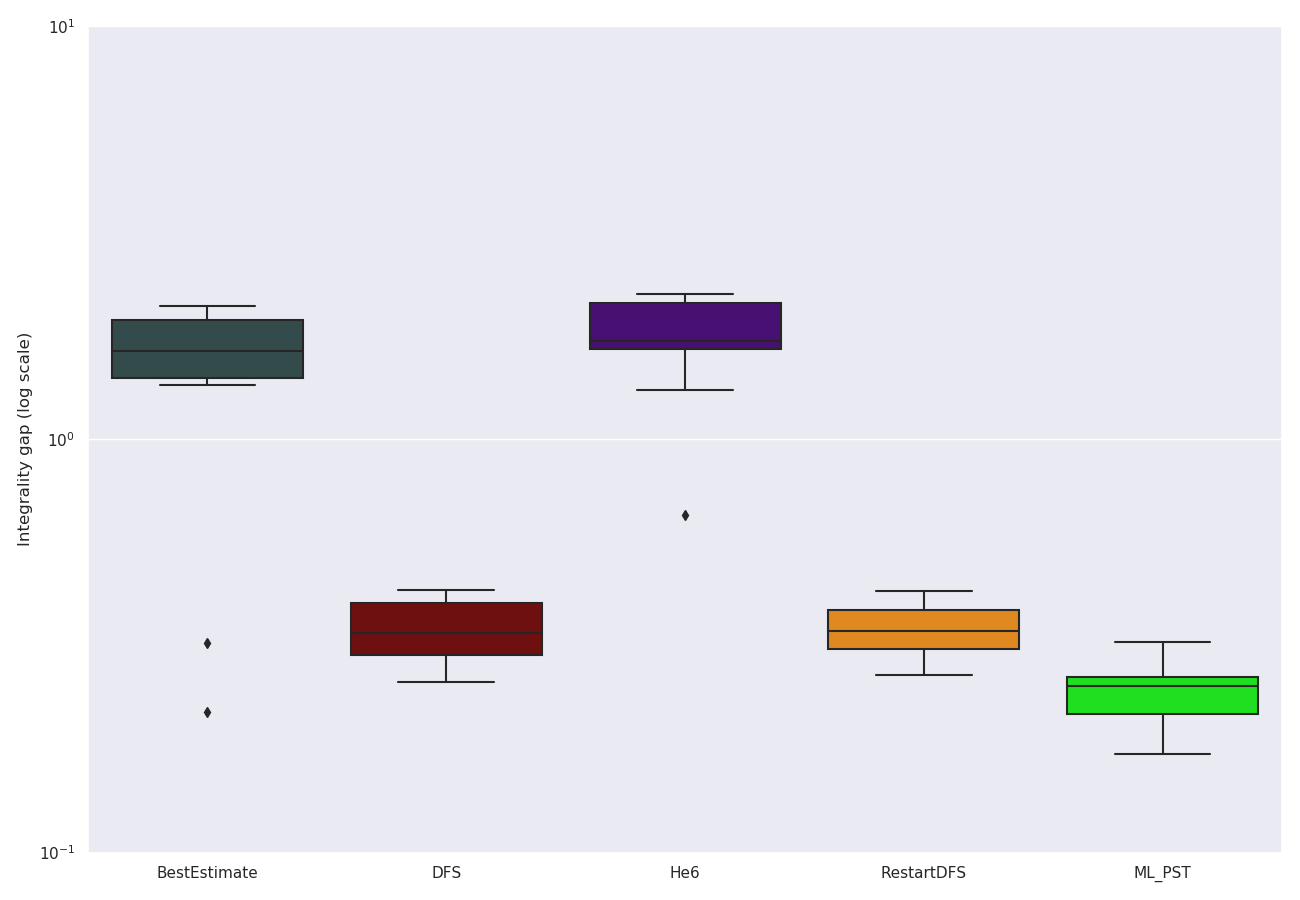}
    \caption{Hard set cover instances: integrality gap of various heuristic strategies.  Note log scale on y-axis. \policy{ML\_PST} outperforms the compared methods.}
    \label{fig:setcover_hard_approx_boxplot}
\end{figure}

\begin{specialtable}[H]
\setlength{\tabcolsep}{8.5mm} 
\caption{Hard set cover instances.
Pair-wise t-tests against \heur{BestEstimate}:
`*' $p < 0.05$, `$\cdot$' $p < 0.1$.}
\label{table:setcover_hard_selection}
\begin{tabular}{llll}\toprule
\textbf{Strategy} &\textbf{Time (s)} &\textbf{Nodes} &\textbf{Integrality Gap} \\\midrule
BestEstimate &\textbf{2256.69} &161438 &\textbf{0.0481} \\
\policy{ML\_PB} &2497.56 &112869 * &0.0714 \\
\policy{ML\_PR} &2531.32 &150055 &0.0763 \\
\policy{ML\_PS} &2279.41 &157367 &0.1152 \\
\policy{ML\_RB} &2441.60 &111629 * &0.0683 \\
\policy{ML\_RR} &2552.68 &152871 &0.0794 \\
\policy{ML\_RS} &2352.93 &153727 &0.1289 $\cdot$ \\
\policy{ML\_SB} &2427.25 &\textbf{109298} * &0.0706 \\
\policy{ML\_SR} &2685.09 &161825 &0.0734 \\
\policy{ML\_SS} &2381.28 &163631 &0.1278 $\cdot$ \\
\bottomrule
\end{tabular}
\end{specialtable}

\vspace{-12pt}

\begin{figure}[H]
    \includegraphics[width=\linewidth]{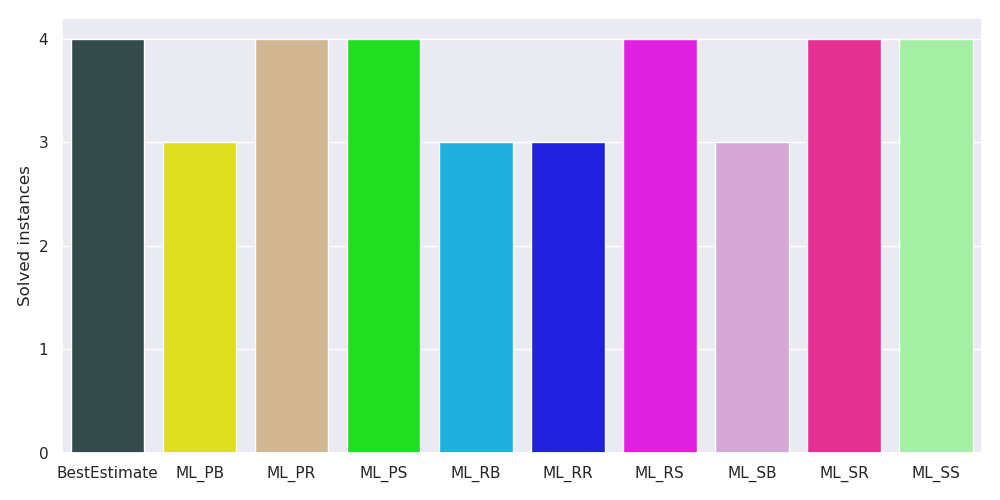}
    \caption{Hard set cover instances: number of solved instances (out of 10) for each node \mbox{selection strategy}.}
    \label{fig:setcover_hard_solved_instances}
\end{figure}

\begin{figure}[H]
    \includegraphics[width=\linewidth,
    clip,trim={0 0 0 1.2cm}]{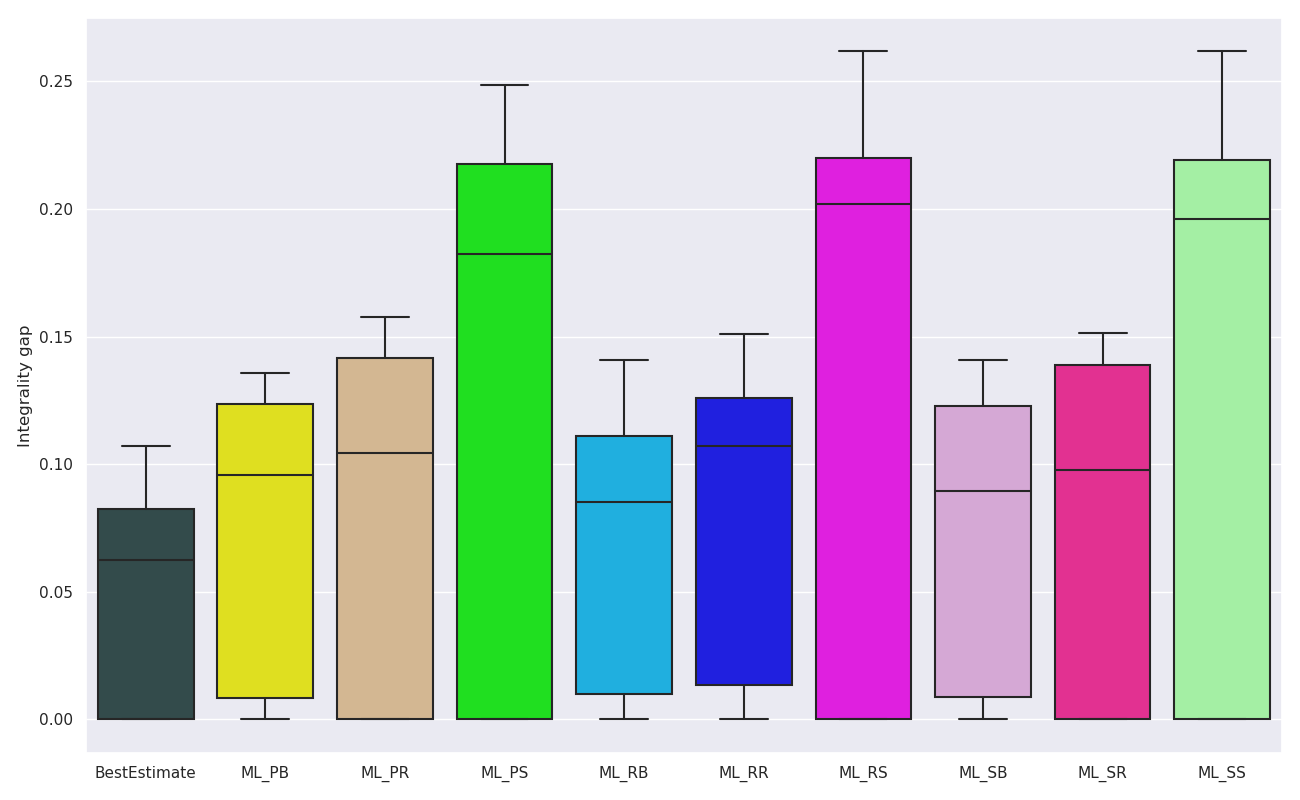}
    \caption{Hard set cover instances: integrality gap of each node selection strategy.}
    \label{fig:setcover_hard_optimality_boxplots}
\end{figure}

\subsection{Success of the Learned Imitation}
\label{sec:results:experiment4}

The results reported so far correspond to the first three experiments: optimality gap and solving time of first solution at a leaf node; total solving time to find an optimal solution and prove its optimality; and optimality gap within a fixed solving time.  \mbox{The results} are on set cover (easy and hard instances), and three further problem classes.

We further analyze the behaviour of the learned policy during the first plunge, \mbox{in detailed} comparison to SCIP's default \heur{BestEstimate} rule. Recall that, when plunging, \heur{BestEstimate} chooses the priority child according to the \emph{PrioChild} property (see Section~\ref{sec:bg}), which is one of the input features to our method.  It is also important to note that, while the learned policy plunges until finding a leaf node, \heur{BestEstimate} may decide to abort the plunge early. This early abort decision is made according a set of pre-established parameters such as  a maximum plunge depth (for more details we refer to the SCIP documentation).

In order to study the trade-off between better feasible solutions and the computational cost of obtaining them, we analyze the attained optimality gap against depth of the final node in the first plunge. This in contrast to our previous experiment, where we considered solving time as the second metric. We compare \heur{BestEstimate} and the learned policy with \policy{on\_both} set to \heur{PrioChild}.  Further, we present experiments per instance, instead of aggregating them.

Figure~\ref{fig:setcover_first_solution_depth} shows the results for the 35 (small) test instances. The effect of \heur{BestEstimate}'s plunge abort becomes apparent. On instances where the first leaf has depth smaller than 14, both policies perform identically. On the contrary, on the remaining instances, \heur{BestEstimate} hits the maximum plunge depth, hence aborting the plunge. These results show that the better optimality gap achieved by the learned policy comes at the cost of processing more nodes, \ie plunging deeper into the tree. We also note that these results were obtained setting on\_both to \heur{PrioChild}, however no node was labelled with {`both'} during our experiments.  In spite of this, the learned policy chose the priority child on 99.6\% of the occasions, demonstrating its strong imitation of that heuristic.

\begin{figure}[H]
    \includegraphics[scale=0.95]{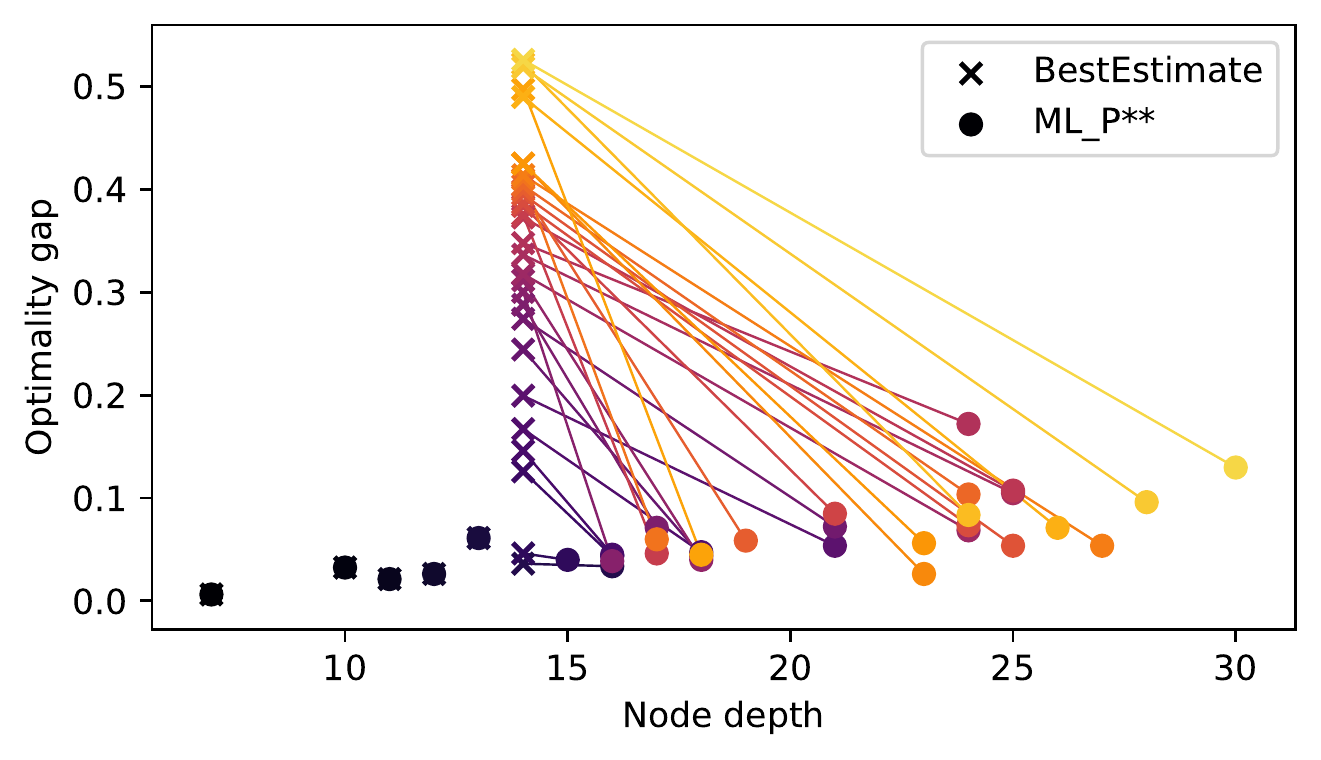}
    \caption{Set cover: node depth vs.\@ optimality gap attained at the final node, after one plunge.  \mbox{Each point-pair} corresponds to one instance: one for \heur{BestEstimate} ($\times$) and one for the learned \mbox{policy ($\bullet$).}}
    \label{fig:setcover_first_solution_depth}
\end{figure}

\section{Discussion and Limitations}
\label{sec:discuss}
This section reflects on our experiments and results, and discusses design decisions and possible alternative to explore.

\subsection{Experiments}
Summarising Section~\ref{sec:results}, we undertook three performance experiments--namely measuring the solving time and optimality gap for the first solution found at a leaf node in the branch and bound tree, measuring the total solving time for an exact solution, and setting a low instance-specific time limit to measure
the optimality gap--and one exploratory experiment.

For the first experiment, the \heur{PrioChild} ML policy ({\policy{ML\_P$\cdot\cdot$}}) performed Pareto-equivalent on four of the five problem sets, performing inferior to the baselines only on maximum independent set instances.
We turned off SCIP's primal heuristics in order to see the ability of the methods to find feasible solutions.

For the second experiment, our method performed better than the baselines in terms of mean solving time for capacitated facility location problem, but worse on the three purely binary problems.  The best ML policy is \policy{ML\_RB}, although \policy{ML\_SB} has only slightly higher time, while exploring fewer nodes.

For the third experiment, we chose \policy{ML\_SRF} on set cover, \policy{ML\_PRF} on maximum independent set, \policy{ML\_SST} on capacitated facility location, and \policy{ML\_PST} on combinatorial auctions and hard set cover instances, in order to measure how well these best ML policies perform against the baselines.  These policies were chosen based on the (lowest) harmonic mean between the solving time and optimality gap.
In four of the five problem sets, our policies had a statistically-significant lower optimality gap than the baselines, \mbox{while on} combinatorial auctions, \policy{ML\_PST} did not perform statistically 
worse than \heur{DFS} and \heur{RestartDFS}.

For Experiments 2 and~3, we kept primary heuristics off, except for hard instances.  \mbox{It would} be interesting to repeat the easy instances with primal heuristics on.  This is because these heuristics improve the primal side of the bounds improvement while the branching rule takes care of dual side--and the node selection must balance both depending on the situation.  The interesting question, which is answered in part by looking at the hard set cover instances, is whether the balance of primal and dual will indicate that the node selection should restrain its diving.

Overall we conclude that the \emph{on\_both} = \heur{Random} configuration of the policy usually performs worse than the other configurations.  \emph{on\_both} $\in$ \{\heur{PrioChild}, \heur{Second}\} both do well.
The policies from both the \emph{on\_leaf} $\in$ \{\heur{Score}, \heur{RestartDFS}\} configurations perform better than those from the \emph{on\_leaf} = \heur{BestEstimate} configuration.
For both \emph{prune\_on\_both} configurations, the policy performed well.  Recall that when \emph{prune\_on\_both} is True, then the search is terminated after the first leaf, saving solving time but resulting in a higher optimality gap.  That both \emph{prune\_on\_both} configurations lead to effective policies means that we offer the user the choice between a lower optimality gap and higher solving time, or the other \mbox{way around}.

Our method is effective when the ML model is able to meaningfully classify optimal child nodes correctly.  By contrast, in the case of the maximum independent set problem, the classification was poor (base acc.: $0.895$, test acc.: $0.899$, gain: $0.004$).  Hence, when the predictive model adds value to the prediction, there is potential for effective decision making using the policy; and contrariwise when it does not.

We note that the feature extraction was the biggest contributor to the overall solving time.  Applying the predictor had a rather small impact.  This means that it is possible to achieve lower solving times by incorporating the entire process in the original C code of SCIP, avoiding the Python interface.
However, Experiment~4 suggests that even a lower overhead in invoking the ML policy will not pay off.

Second, instead of actually pruning nodes, one could only assign low priorities to children considered inferior, such that they are never selected after a backtrack.  This would automatically make the solver with the learned policy exact since it would never commit to pruning.  However, the objective for our work was to explore learning approximate node selection policies.  Second, node pruning has the advantage that the search does not visit the subtree of that node at all.  Third, by pruning we follow the direction of  He et al. \citep{He2014:learning} and can compare directly to the previous state-of-art in the literature.
Indeed, 
as seen above,
our approach easily outperforms 
that of He et al. \citep{He2014:learning}, in both their original implementation and a re-implementation in SCIP~6.

\subsection{Using Best $k$ Solutions}
As discussed earlier, we set $k=10$ for all our experiments bar one, with the hypotheses that, in general, higher values of $k$ allows the ML model better potential to generalise.  \mbox{We found} that lower $k$ (below $10$) gave inferior results in initial experiments.  There is potential to vary $k$ according to the problem class being solved.  In particular,  the distribution of nodes labelled `L', `R' and `B'.  If $k$ is too high, it could be the number of `B's become too high, leading to a class imbalance in training the ML model.  On the other hand, another artefact of $k$ being too low can be class imbalance, as we saw for maximum independent set instances.

We make three further remarks.  First, we rank solutions using SCIP's solution ranking which is based on the optimality gap.  It could be interesting to look also at the depth of the solutions in the tree, especially for the first setting of the heuristic policy use.  Second, the $k$ best solutions can vary in their objective value, possibly by relatively large amounts.  An alternative to using any number of best solutions is to select an optimality gap bound.  However, such an approach is more complicated than using the top-$k$.
Third, in this article we trained on instances where the optimum was known.  In the case it is not known, \mbox{one could} choose the best $k$ solutions in terms of their integrality gap.

In Section~\ref{sec:conc} we discusses additional possible further work for choosing $k$.

\subsection{Heuristic and Exact Settings}
Recall that our learned policies were studied in two settings.
In the first setting, \mbox{the learned} policy is used as a heuristic method.
This is akin to a diving heuristic, with the difference that it uses the default branching rule as a variable fixing strategy.
In more detail, this diving heuristic uses the branching rule to decide on which variable the disjunction will be made and then uses the child selector to choose the value to which the variable will be fixed (\ie 0 or 1).  In MIP branch-and-bound trees it is known that good branching rules are not good diving rules {(attributed to T.~Berthold).}  This is because the branching rule explicitly tries to balance the quality of the two children, in order to focus mostly on improving the lower bound, whereas a diving rule will try to construct an unbalanced (one-sided) tree.
It remains to compare the learned policy against actual diving heuristics.  We hypothesise that it will be inferior in terms of total time, because our heuristic uses the default branching rule as a variable fixing strategy.

In the second setting--the exact setting--the learned policy is used as a child selector during plunging.  The potential of learning here is to augment node selection rules, \mbox{such that} the child selection process is better informed than the simple heuristics that solvers typically use.  Recall that in the case of SCIP, the heuristic is \heur{BestEstimate} with plunging.
Indeed, our fourth experiment showed that, when used as a child selector, the learned policy acts almost exactly like SCIP's \emph{PrioChild} rule.  The main difference comes from the fact that SCIP's plunging has an abort mechanism.  The policy could not learn this because policy learns to select a child, and then the node selection rule dives using this policy until no children are left.  That the policy acts like \emph{PrioChild} is no surprise given that this is the rule that was used to generate samples and it is also one of the features fed to the learner. 
Table~\label{tab:setcover_selection} shows that, in terms of solving time, the learned policy is not better than \heur{BestEstimate} with plunging and its abort mechanism.

We conclude that that there is limited potential for improvement by selecting the best child during plunging.  If the branching rule is working well, the two children should be quite balanced.  By contrast, an interesting question is choosing a good node \emph{after} plunging stops.

\section{Related Work}
\label{sec:rw}
We position our work in the literature on using machine learning in MIP branching, node selection, and other aspects of MIP solving.

\subsection{Branching}
\label{section:branching}

Deciding on what variable to branch on in the branch and bound process is called branching, as was introduced in the main text of the paper.  Good branching techniques make it possible to reduce the tree size, resulting in fast solving times.  A survey on branching, and the use of learning to improve it, is by \citet{Lodi17:learn-branch}.

Strong branching \citep{applegate1995finding} is a popular branching strategy, among other strategies such as most-infeasible branching, pseudo-cost branching \citep{DBLP:journals/mp/BenichouGGHRV71}, reliability branching \citep{achterberg2005branching}--used as the default in SCIP--and hybrid branching \citep{achterberg2009hybrid}. 
Strong branching creates the smallest trees, as Achterberg  et al. \citep{achterberg2005branching} reported that strong branching required around 20 times fewer nodes to solve a problem than most infeasible branching and around 10 times fewer nodes than pseudo-cost branching.  However, strong branching is the most expensive to calculate, because two LP-relaxations are solved for every variable to assign scores. 

Nonetheless, exact scores are not required to find the best variable to branch on.  Therefore, it is interesting to approximate the score of strong branching, which can be done using machine learning.
Alvarez  et al. \citep{alvarez2014supervised} was the first to use supervised learning to learn a strong branching model. 
The features they used to train the ML model consist of static problem features, dynamic problem features and dynamic optimisation features. The static problem features derive from $c$, $A$ and $b$ as stated in Equation \ref{MILP-formulation}. The dynamic problem features derive from the solution $\hat{x}$ of the current node in the branch and bound tree and the dynamic optimisation features derive from statistics of the current variable. They used the Extremely Randomized Trees (ExtraTree) classifier \citep{geurts2006extremely}.  The results show that supervised learning successfully imitated strong branching, being 9\% off relative to gap size, but 85\% faster to calculate.
Although strong branching was successfully imitated, it was still behind reliability branching in terms of gap size and runtime.

Khalil et al. \citep{khalil2016learning} extended Alvarez  et al. \citep{alvarez2014supervised} work by adding new features to the machine learning model and by learning a pairwise ranking function instead of a scoring function. The ranking function they used is a ranking variant of Support Vector Machine (SVM) classifier \citep{joachims2006training}. 
Their algorithm solved 70\% more hard problems (over 500,000 nodes, cut-off time 5 h) than strong branching alone. However, the time spent per node (18 ms) is higher than pseudo-cost branching (10 ms) and combining strong branching with pseudo-cost branching (15 ms). This is due to calculating the large number of features on every node.

To overcome complex feature calculation, Gasse et al. \citep{DBLP:conf/nips/GasseCFC019} proposes features based on the bipartite graph structure of a general MILP problem. The graph structure is the same for every LP relaxation in the branch-and-bound tree, which reduces the feature calculation cost.  They use a graph convolutional neural network (GCNN) to train and output a policy, which decides what variable to branch on.  Furthermore, they used cutting planes on the root node to restrict the solution space.  Their GCNN model performs better than both Alvarez et al.  \citep{alvarez2014supervised} and Khalil et al.  \citep{khalil2016learning} for generalising branching, using few demonstration examples for the set covering, capacitated facility location, and combinatorial auction problems.
Moreover, GCNN solved the combinatorial auction problem 75\% faster than the method of Alvarez et al.  \citep{alvarez2014supervised} and 
70\% faster than the method of Khalil et al.  \citep{khalil2016learning}, both for hard problems (1,500 auctions).
Seeing their success, we adopt the same variable features as \citet{DBLP:conf/nips/GasseCFC019}.

Other recent works learning branching rules are \citet{DBLP:journals/corr/abs-2002-05120,DBLP:conf/nips/GuptaGKM0B20,Yang2020:learn-gsb}.

\citet{DBLP:journals/corr/abs-2012-13349} develop a combination of branching rule and primal heuristic.
The authors apply imitation learning to acquire a MIP brancher, based on full strong branching \citep{achterberg2005branching}. 
Separately, the authors apply generative modelling and supervised learning to acquire a `diving rule'.
Together, SCIP using the learned brancher and learned diving rule outperforms default SCIP by an order of magnitude or more in terms of the primal integral.  It is noteworthy that \citet{DBLP:journals/corr/abs-2012-13349} train and test on relatively large MIP instances and use extensive amount of parallelised GPU computation in order to train their neural \mbox{network models.}

\subsection{Node Selection and Pruning Policy}

While learning to branch has been studied quite extensively, learning to select and prune nodes has received insufficient attention in the literature.

He et al. \citep{He2014:learning} used machine learning to imitate good node selection and pruning policies.  The method of data collection in that work is by first solving a problem and provide its solution to the solver.  Afterwards, the problem is solved again, but now that the solver knows the solution, it will take a shorter path to the solution.  The features for learning the node selection policy are derived from the nodes in this path and the features for the node pruning policy are derived from the nodes that were not explored further. This was done for a limited amount of problems
as the demonstrations.

He et al. \citep{He2014:learning} trained their machine learning algorithm on four datasets, called MIK, Regions, Hybrid and CORLAT. They were able to achieve prune rates of 0.48, 0.55, 0.02 and 0.24 for each dataset respectively. Prune rate shows the amount of nodes that did not have to be explored further relative to the total amount of nodes seen.  Their solving time reached a speedup of 4.69, 2.30, 1.15 and 1.63 compared to a baseline SCIP version~3 heuristic respectively for each dataset.  Note that the lowest speedup seems to correlate with a low prune rate.

Our work differs from He et al. \citep{He2014:learning} by constraining the node selection space to direct children only at non-leaf nodes.  Second, we use the top $k$ solutions to sample state-action pairs. By using more than one solution, we can create additional state-action pairs from which the neural network can learn and create a predictive model.  Third, we take advantage of branched variable features, obtained from Gasse et al.  \citep{DBLP:conf/nips/GasseCFC019}. 
As seen in Section~\ref{sec:results}, our approach easily outperforms that of He et al. \citep{He2014:learning}, in both their original implementation and a re-implementation in SCIP 6.

Other recent works learning decisions related to node selection and pruning focus on learning primal heuristics, sometimes for a specific problem class \citep{DBLP:conf/aaai/DingZSLWXS20,DBLP:journals/corr/abs-2012-13349,DBLP:journals/eor/BengioLP21}.

\section{Conclusions}
\label{sec:conc}
This article shows that approximate solving of mixed integer programs can be achieved by a node selection policy obtained with offline imitation learning.  Node selection is important because through it, the solver balances between exploring search nodes with good lower bounds (crucial for proving global optimality), and finding feasible solutions fast.
In contrast to previous work using imitation learning, our policy is focused on learning to choose which of its children it should select.  We apply the policy within the popular open-source solver SCIP, in exact and heuristic settings.

Empirical results on five MIP datasets indicate that our node selector leads to solutions more quickly than the state-of-the-art in the literature \citep{He2014:learning}.  While our node selector is not as fast on average as the highly optimised state-of-practice in SCIP in terms of solving time on exact solutions, our heuristic policies have a consistently better optimality gap than all baselines if the accuracy of the predictive model is sufficient.
Second, the results also indicate that our heuristic method finds better solutions within a given time limit than all baselines in the majority of the problem classes examined.
Third, the results show
that learned policies can be Pareto-equivalent or superior to state-of-practice MIP node selection heuristics.
However, these results come with a caveat, summarized below.

This work adds to the body of literature that demonstrates how ML
can benefit generic constraint optimisation problem solvers.
In MIP terminology, our learned policy constitutes a diving rule, focusing on finding a good integer feasible solution.  The performance on non-binary problem classes like capacitated facility location is particularly noteworthy.  This is because, unlike purely binary problems, for non-binary instances, MIP primal heuristics struggle to obtain decent primal bounds \citep{achterberg2008constraint}.  By contrast, in general for binary instances, the greater challenge is to close the dual bound, and our learned policy also performs well here.

The results are explained by the conclusion that the learned policies have imitated SCIP brancher's preferred rule for node selection, but without the latter's early plunge abort.  This is a success for imitation learning, but does not overall improve the best state-of-practice baseline from which it has learned.  Pareto-efficiency between total solving time and integrality gap is is important, yet total solving time and the primal integral quality are the most crucial in practice.  However, an interesting question to pursue is learning to choose a good node \emph{after} plunging stops.

Despite the clear improvements over the literature, then, this kind of MIP child selector is better seen in a broader approach to using learning in MIP branch-and-bound tree decisions \citep{DBLP:journals/corr/abs-2012-13349}.  While forms of supervised learning find success \citep{DBLP:journals/eor/BengioLP21}, reinforcement learning is also interesting \citep{DBLP:conf/nips/GasseCFC019}, and could be used for node selection.

Nonetheless, for future work on node selection by imitation learning, more study could be undertaken for choosing the meta-parameter $k$.  Values too low add only few state-action pairs, which naturally degrades the predictive power of neural networks.  \mbox{On the} other hand, values too high add noise, as paths to bad solutions add state-action pairs that are not useful.
An interesting direction is to exploit an oracle (solver) to decide whether a node is `good' or `bad', \eg if the node falls onto a path of a solution within say 5\% of the optimum value.  This more expensive data collection method might eliminate choosing a specific $k$.

The way in which the ML policies were trained can be explored further.  For instance, one could consider the two components of \emph{PrioChild} separately, or could train on standard SCIP with and without early plunge abort.  Comparing learned policies with and without primal heuristics and with and without pre-solve also has scientific value.

Lastly, Section~\ref{sec:method} explained how during pre-processing certain features are removed which are constant throughout the entire dataset.  This has the consequence of a different number of input units in the neural network architecture for every problem.
Moreover, future
work could include a method to unify a ML model that is effective for all problem classes.  This would make ML-based node selection a more accessible feature for current MIP solvers such as SCIP; another promising direction in this regard is experimentation via the `gym' of Prouvost et al. \citep{DBLP:journals/corr/abs-2011-06069}.

\authorcontributions{
Conceptualization, K.Y.\@ and N.Y.-S.; methodology, K.Y.; software, K.Y.; analysis, K.Y.; writing---original draft preparation, K.Y.; writing---review and editing, N.Y.-S.; visualization, K.Y.; supervision, N.Y.-S.; project administration, N.Y.-S. All authors have read and agreed to the published version of the manuscript. Note this is an authors' preprint, not the publisher's version of the article.}

\funding{This research was partially supported by TAILOR, a project funded by EU Horizon 2020 research and innovation programme under grant 952215, and by the Dutch Research Council (NWO) Groot project OPTIMAL.}

\institutionalreview{Not applicable.}
\informedconsent{Not applicable.}

\dataavailability{
Source code for the methods of this article available is available at \url{www.doi.org/10.4121/14054330}.  Problem instance were obtained from the generator of  Gasse et al.\@ \citep{DBLP:conf/nips/GasseCFC019} with default settings.}

\acknowledgments{%
Thanks to Lara Scavuzzo Monta{\~{n}}a for a number of contributions.  Thanks to Robbert Eggermont for computational support.  We thank  {the anonymous reviewers of this article, and} those who commented on the arXiv preprint of this work since July~2020.}

\conflictsofinterest{The authors declare no conflict of interest.}

\appendixtitles{yes}
\appendixstart
\appendix

\section{Results on Three Further Problems Sets}

Besides easy and hard set cover instances, as reported in Section~\ref{sec:results}, we studied our ML-based node selection and pruning policies on three further standard datasets.  These are maximum independent set, capacitated facility location and combinatorial auctions.
Instances were obtained from the generator of  Gasse et al.\@ \citep{DBLP:conf/nips/GasseCFC019} with default settings.

\subsection{Maximum Independent Set}
\label{section:indset}

These instances consist of 1000 variables and around 4000 constraints forming a pure binary maximisation problem.  For this particular problem, we noticed that for $k = 10$, \mbox{the class} imbalance was significant.  To mitigate this imbalance, we increased the value $k$ to 40 (see the discussion in Section~\ref{sec:discuss}).  For $k = 10$, we sampled 29,801 state-action pairs on the training instances, 5820 on the validation instances and 4,639 on the testing instances. The class distribution is: (Left: 92\%, Right: 4\%, Both: 4\%).
For $k = 40$, we sampled 82,986 state-action pairs on the training instances, 14,460 on the validation instances and 14,273 on the testing instances. 
The class distribution is: (Left: 89\%, Right: 4\%, Both: 7\%).

Both the $k = 10$ and $k = 40$ models achieve a testing accuracy that is very close to the baseline accuracy, which results in a model that is not able to generalise.
See \mbox{Table~\ref{table:indset_selection}} for the average solving time and explored nodes of various node selection strategies.
\mbox{We conducted} a pair-wise t-test between the mean solving time of \heur{BestEstimate} and the mean solving time of the other policies.  We can reject the null hypothesis of equal means with \emph{p}-value below 0.1 for all our ML policies.
We have also conducted a pair-wise t-test between the mean number of explored nodes of \heur{BestEstimate} and the mean number of explored nodes of the other policies. We can reject the null hypothesis of equal means with \emph{p}-value below 0.1 for 12 of 18 of our ML policies.

\begin{specialtable}[H]
\setlength{\tabcolsep}{11mm} 

\caption{Maximum independent set instances: average solving time and explored nodes for various node selection strategies. Pair-wise t-tests against \heur{BestEstimate}: `***' $p < 0.001$, `**' $p < 0.01$, `*' $p < 0.05$, `$\cdot$' $p < 0.1$}
\label{table:indset_selection}
\begin{tabular}{llll}\toprule
\textbf{Strategy} &\textbf{Solving Time (s) }&\textbf{Explored Nodes} \\\midrule
BestEstimate &158.42 &6344 \\
DFS &\textbf{155.27} &\textbf{6340} \\
He (S) &394.56 ** &30965 ** \\
He6 (S) &204.68 &7992\\
\policy{ML\_PB (10)} &260.34 ** &10029 * \\
\policy{ML\_PB (40)} &227.45 * &8100 \\
\policy{ML\_PR (10)} &255.49 ** &10191 * \\
\policy{ML\_PR (40)} &222.59 * &8581 \\
\policy{ML\_PS (10)} &245.92 * &10184 $\cdot$ \\
\policy{ML\_PS (40)} &207.28 $\cdot$ &7878 \\
\policy{ML\_RB (10)} &274.17 ** &10296 * \\
\policy{ML\_RB (40)} &222.80 * &8006 \\
\policy{ML\_RR (10)} &244.01 ** &9654 * \\
\policy{ML\_RR (40)} &213.81 $\cdot$ &8196 \\
\policy{ML\_RS (10)} &232.67 * &9582 $\cdot$ \\
\policy{ML\_RS (40)} &207.71 $\cdot$ &8137 \\
\policy{ML\_SB (10)} &285.21 ** &10661 * \\
\policy{ML\_SB (40)} &283.02 ** &10491 * \\
\policy{ML\_SR (10)} &252.59 ** &9936 * \\
\policy{ML\_SR (40)} &258.68 ** &10120 * \\
\policy{ML\_SS (10)} &242.37 * &9921 $\cdot$ \\
\policy{ML\_SS (40)} &274.73 ** &11251 * \\
RestartDFS &183.24 &7854 \\
\bottomrule
\end{tabular}

\end{specialtable}

For both $k = 10$ and $k = 40$, \heur{DFS} achieved the lowest average solving time on node selection at 155.3 s, while the $k = 10$ \policy{ML\_RS} model achieved 232.7 s and the $k = 40$ \policy{ML\_PS} model an average solving time of 207.3 s.

Figure~\ref{fig:indset_10_40_pareto} shows that the first solution quality and solving time of ML policies are all near each other and dominated by \heur{RestartDFS} and \heur{DFS}.  Note that in the plot the suffix (`\policy{**}') is replaced by the value of $k$.
Table~\ref{table:policy_eval_indset_40} 
examines how the node pruner compares to the baselines, when the baselines have a set time limit.  In this case,  \policy{ML\_PRF} has the lowest harmonic mean between the average solving time and average optimality gap of all ML policies. The ML policy has a higher average optimality gap than the baselines for this problem.  
We conducted a pairwise t-test between the mean optimality gap of our best ML policy and the mean optimality gap of each baseline.  We can reject the null hypothesis of equal means with \emph{p}-value below 0.05 for all baselines, except BestEstimate (\emph{p}-value: 0.334).
The initial optimality gap obtained by the solver before branch-and-bound is $0.999$.
This shows that He6 policy prunes aggressively at the start, because the average optimality gap obtained by He6 is similar to initial optimality gap.
The other policies find significantly better solutions.

\begin{figure}[H]
        \includegraphics[width=0.95\linewidth,clip,trim={0 0 0 1.2cm}]{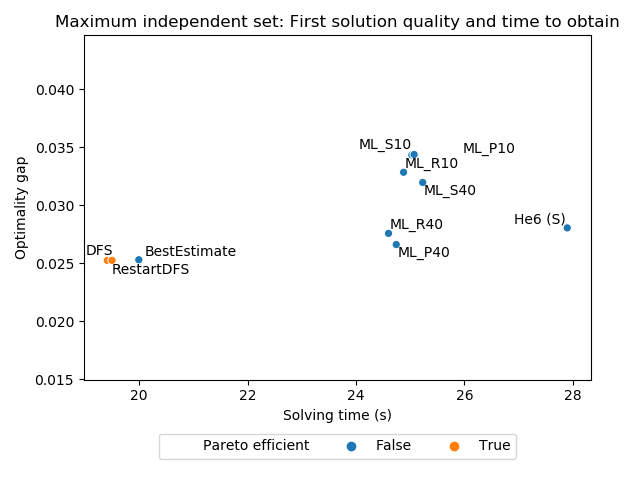}
        \caption{Maximum independent set: average solving time against the average optimality gap of the first solution found at a leaf node.}
        \label{fig:indset_10_40_pareto}
        \end{figure}

\vspace{-6pt}
        \begin{specialtable}[h]
        \setlength{\tabcolsep}{23mm} 
\caption{Maximum independent set: model ($k = 40$) with \textit{on\_both} = PrioChild, \textit{on\_leaf} = RestartDFS and \textit{prune\_on\_both} = False against baselines, with equal time limits for each problem. The initial optimality gap obtained by the solver before branch-and-bound is $0.999$. Pairwise t-tests against the ML policy: `***' $p < 0.001$, `**' $p < 0.01$, `*' $p < 0.05$, `$\cdot$' $p < 0.1$.}
\label{table:policy_eval_indset_40}
\begin{tabular}{ll}\toprule
\textbf{Strategy} &\textbf{Optimality Gap} \\\midrule
BestEstimate &0.0174 \\
DFS     &\textbf{0.0134} * \\
He6 (B) &0.9930 *** \\
He6 (P) &0.9902 *** \\
\policy{ML\_PRF} &0.0211 \\
RestartDFS &0.0134 * \\
\bottomrule
\end{tabular}

\end{specialtable}
\vspace{-12pt}

\subsection{Capacitated Facility Location}
These instances consist of 150 binary variables, 22,500 continuous variables and 300 constraints, forming a mixed-integer minimisation problem.  We sampled 17,266 state-action pairs on the training instances, 3531 on the validation instances and 3,431 on the testing instances.  The model achieves a testing accuracy of 90.1\%, with a baseline of 73.1\%.  

See Table~\ref{table:facilities_selection} for the average solving time and explored nodes of various node selection strategies.  \policy{ML\_RB} achieves the lowest average solving time at $111.7$ seconds and \policy{ML\_SS} the lowest average explored nodes at $1099$.
We conducted a pair-wise t-test between the mean solving time of \heur{BestEstimate} and the mean solving time of the other policies.  We can not reject the null hypothesis of equal means with \emph{p}-value below 0.1 for any our ML policies (lowest observed \emph{p}-value: 0.14).
We have also conducted a pair-wise t-test between the mean number of explored nodes of \heur{BestEstimate} and the mean number of explored nodes of the other policies. We can not reject the null hypothesis of equal means with \emph{p}-value below 0.1 for any our ML policies (lowest observed \emph{p}-value: 0.18).

\begin{specialtable}[H] 
 \setlength{\tabcolsep}{11mm} 
\caption{Capacitated facility location instances: average solving time and explored nodes for various node selection strategies. Pair-wise t-tests against \heur{BestEstimate}: `***' $p < 0.001$, `**' $p < 0.01$, `*' $p < 0.05$, `$\cdot$' $p < 0.1$}
\label{table:facilities_selection}
\begin{tabular}{lll}\toprule
\textbf{Strategy} &\textbf{Solving Time (s)} &\textbf{Explored Nodes }\\\midrule
BestEstimate &122.79 &1674 \\
DFS &690.40 *** &17155 *** \\
He (S) &1754.10 *** &5733 *** \\
He6 (S) &373.32 *** &6799 *** \\
\policy{ML\_PB} &114.41 &1189 \\
\policy{ML\_PR} &148.21 &1740 \\
\policy{ML\_PS} &118.36 &1207 \\
\policy{ML\_RB} &\textbf{111.67} &1163 \\
\policy{ML\_RR} &147.85 &1773 \\
\policy{ML\_RS} &125.99 &1344 \\
\policy{ML\_SB} &111.69 &1109 \\
\policy{ML\_SR} &133.68 &1462 \\
\policy{ML\_SS} &115.39 &\textbf{1099} \\
RestartDFS &444.36 *** &9977 *** \\
\bottomrule
\end{tabular}

\end{specialtable}

Figure~\ref{fig:facilities_pareto} shows the average solving time against the average optimality gap of the first solution obtained by the baselines and ML policies at a leaf node. We see here that the ML policies are clustered and obtain a lower optimality gap than the baselines. \mbox{Note that} \heur{RestartDFS} and \heur{DFS} have a very similar optimality gap and solving time, so they are stacked on top of each other.  See Table~\ref{table:policy_eval_facilities} 
for the optimality gap of the baselines using a time limit for each instance.  In this case, \policy{ML\_SST} has the lowest harmonic mean between the average solving time and average optimality gap of all ML policies.  \policy{ML\_SST} also achieves a significantly lower average optimality gap than the baselines.  
We conducted a pairwise t-test between the mean optimality gap of our best ML policy and the mean optimality gap of each baseline.  We can reject the null hypothesis of equal means with \emph{p}-value below 0.001 for all baselines.
The initial optimality gap obtained by the solver before branch-and-bound is $0.325$.
All policies find a significantly better solution than the first found feasible solution.

\begin{figure}[H]
        \includegraphics[width=0.95\linewidth,clip,trim={0 0 0 1.2cm}]{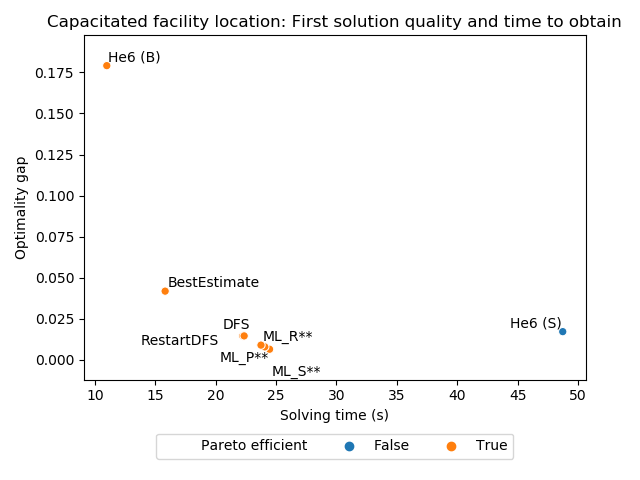}
        \caption{Capacitated facility location: average solving time against the average optimality gap of the first solution found at a leaf node.}
        \label{fig:facilities_pareto}
    \end{figure}
    
\vspace{-6pt}
      \begin{specialtable}[h]
       \setlength{\tabcolsep}{23mm} 
\caption{Capacitated facility location: model with \textit{on\_both} = Second, \textit{on\_leaf} = Score and \textit{prune\_on\_both} = True against baselines, with equal time limits for each problem. The initial optimality gap obtained by the solver before branch-and-bound is $0.325$. Pairwise t-tests against the ML policy: `***' $p < 0.001$, `**' $p < 0.01$, `*' $p < 0.05$, `$\cdot$' $p < 0.1$.}
\label{table:policy_eval_facilities}
\begin{tabular}{ll}\toprule
\textbf{Strategy} &\textbf{Optimality Gap} \\\midrule
BestEstimate &0.0821 *** \\
DFS     &0.0619 *** \\
He6 (B) &0.1516 *** \\
He6 (P) &0.1503 *** \\
\policy{ML\_SST} &\textbf{0.0065} \\
RestartDFS &0.0590 *** \\
\bottomrule
\end{tabular}
\end{specialtable}

\vspace{-12pt}

\subsection{Combinatorial Auctions}
These instances consist of 1200 variables and around 475 constraints forming a pure binary maximisation problem.  We sampled 13,554 state-action pairs on the training instances, 2389 on the validation instances and 2170 on the testing instances.  The model achieves a testing accuracy of 71.7\%, with a baseline of 57.0\%.

See Table~\ref{table:cauctions_selection} for the average solving time and explored nodes of various node selection strategies.  \heur{BestEstimate} achieves the lowest average solving time at $19.7$ s.
\mbox{We conducted} a pair-wise t-test between the mean solving time of \heur{BestEstimate} and the mean solving time of the other policies.  We can reject the null hypothesis of equal means with \emph{p}-value below 0.05 for all our ML policies.
We have also conducted a pair-wise t-test between the mean number of explored nodes of \heur{BestEstimate} and the mean number of explored nodes of the other policies. We can reject the null hypothesis of equal means with \emph{p}-value below 0.1 for \policy{ML\_PS} and \policy{ML\_RS}.

\begin{specialtable}[H] 
\setlength{\tabcolsep}{11mm} 
\caption{Combinatorial auctions instances: average solving time and explored nodes for various node selection strategies. Pair-wise t-tests against \heur{BestEstimate}: `***' $p < 0.001$, `**' $p < 0.01$, `*' $p < 0.05$, `$\cdot$' $p < 0.1$.}
\label{table:cauctions_selection}
\begin{tabular}{lll}\toprule
\textbf{Strategy} &\textbf{Solving Time (s)} &\textbf{Explored Nodes} \\\midrule
BestEstimate &\textbf{19.68} &\textbf{3489} \\
DFS &23.48 &4490 \\
He (S) &40.11 *** &4519 \\
He6 (S) &30.44 ** &6358 * \\
\policy{ML\_PB} &29.77 * &4288 \\
\policy{ML\_PR} &29.82 ** &4419 \\
\policy{ML\_PS} &32.51 ** &4811 $\cdot$ \\
\policy{ML\_RB} &30.08 * &4494 \\
\policy{ML\_RR} &30.89 ** &4715 \\
\policy{ML\_RS} &32.68 ** &5190 $\cdot$ \\
\policy{ML\_SB} &28.94 * &4187 \\
\policy{ML\_SR} &29.83 ** &4458 \\
\policy{ML\_SS} &30.52 ** &4567 \\
RestartDFS &22.35 &4299 \\
\bottomrule
\end{tabular}

\end{specialtable}

Figure~\ref{fig:cauctions_pareto} shows the average solving time against the average optimality gap of the first solution obtained by the baselines and ML policies at a leaf node. We see here that the ML policies are not as clustered.  The \policy{ML\_P**} strategy is the only strategy that delivers Pareto efficient result, having a both a lower optimality gap and a lower solving time.  \mbox{Note that} \heur{BestEstimate}, \heur{RestartDFS} and \heur{DFS} have a very similar optimality gap and solving time, so they are stacked on top of each other.  See Table~\ref{table:policy_eval_cauctions} 
for the optimality gap of the baselines using a time limit for each instance. In this case, \policy{ML\_PST} has the lowest harmonic mean between the average solving time and average optimality gap of all ML policies.  \policy{ML\_PST} achieves a very similar optimality gap compared to the baselines.  
We conducted a pairwise t-test between the mean optimality gap of our best ML policy and the mean optimality gap of each baseline.  We can reject the null hypothesis of equal means with \emph{p}-value below 0.05 for \heur{BestEstimate} and \heur{He6}, but not \heur{DFS} (\emph{p}-value: 0.94) and \heur{RestartDFS} (\emph{p}-value: 0.98).
The initial optimality gap obtained by the solver before branch-and-bound is $0.914$.
All policies find a significantly better solution than the first found feasible solution.

\begin{figure}[H]
        \includegraphics[width=0.95\linewidth,clip,trim={0 0 0 1.2cm}]{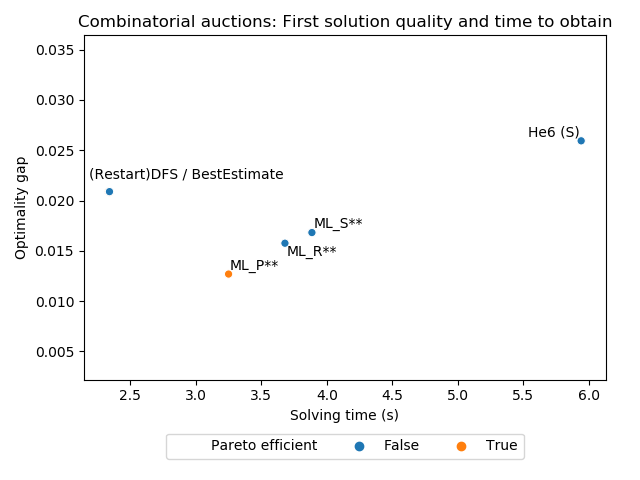}
        \caption{Combinatorial auctions: average solving time against the average optimality gap of the first solution found at a leaf node.}
        \label{fig:cauctions_pareto}
    \end{figure}
    \vspace{-6pt}
 \begin{specialtable}[h]
 \setlength{\tabcolsep}{23mm} 
\caption{Combinatorial auctions: model with \textit{on\_both} = PrioChild, \textit{on\_leaf} = Score and \textit{prune\_on\_both} = True against baselines, with equal time limits for each problem. The initial optimality gap obtained by the solver before branch-and-bound is $0.914$. Pairwise t-tests against the ML policy: `***' $p < 0.001$, `**' $p < 0.01$, `*' $p < 0.05$, `$\cdot$' $p < 0.1$.}
\label{table:policy_eval_cauctions}
\begin{tabular}{ll}\toprule
\textbf{Strategy} & \textbf{Optimality Gap }\\\midrule
BestEstimate &0.0174 * \\
DFS     &\textbf{0.0126} \\
He6 (B) &0.4678 *** \\
He6 (P) &0.2873 *** \\
\policy{ML\_PST} &0.0127 \\
RestartDFS &0.0126 \\
\bottomrule
\end{tabular}
\end{specialtable}

\end{paracol}
%
\reftitle{References}

\end{document}